\documentclass[journal]{IEEEtran}
\usepackage{lipsum}
\usepackage{mathtools}
\usepackage{cuted}
\usepackage{caption}
\usepackage{subcaption}
\usepackage{epsfig} % for postscript graphics files
\usepackage{times} % assumes new font selection scheme installed
\usepackage{amsmath} % assumes amsmath package installed

\usepackage{amsthm}
\usepackage{amssymb}  % assumes amsmath package installed
\usepackage{cite}
\usepackage{psfrag}
\usepackage{cancel}
\usepackage{tikz}
\usepackage{accents}

\usepackage{color}
\usepackage{balance}
\usepackage{array}
\usepackage{breqn}
\usepackage{multirow}

\theoremstyle{remark}
\newtheorem{remark}{Remark}
\renewcommand\vec\boldsymbol

\usepackage{subcaption}       % the caption 
\usepackage{caption}
\usepackage{booktabs}                     % added for  3 line table
\renewcommand{\thetable}{\Roman{table}}
\renewcommand{\tablename}{TABLE}

\title{
	Fully distributed cooperation for networked uncertain mobile manipulators
}

\author{% <-this % stops a space 
	Yi Ren, Sandra Hirche, \emph{Senior Member, IEEE}
    \thanks{This work is supported by the joint Sino-German Research Project
”Control and optimization for event-triggered networked autonomous 
multi-agent systems (COVEMAS)", which is funded through the German 
Research Foundation (DFG) and the National Science Foundation China (NSFC)."}
	\thanks{Yi Ren and Sandra Hirche are with the Chair of Information-Oriented Control, Department of Electrical and Computer Engineering, Technische Universit\"at M\"unchen, Munich, Germany. (e-mail: yi.ren@tum.de; hirche@tum.de).}
}

\begin{document}

	\maketitle
	
\IEEEpubid{\begin{minipage}{\textwidth}\ \\[12pt] \centering
  \copyright~2020 IEEE.  Personal use of this material is permitted.  Permission from IEEE must be obtained for all other uses, in any current or future media, including reprinting/republishing this material for advertising or promotional purposes, creating new collective works, for resale or redistribution to servers or lists, or reuse of any copyrighted component of this work in other works.
\end{minipage}} 	
	
	%%%%%%%%%%%%%%%%%%%%%%%%%%%%%%%%%%%%%%%%%%%%%%%%%%%%%%%%%%%%%%%%%%%%%%%%%%%%%%%%
	%%%%%%%%%%%%%%%%%%%%%%%%%%%%%%%%%%%%%%%%%%%%%%%%%%%%%%%%%%%%%%%%%%%%%%%%%%%%%%%%
	\begin{abstract}
This paper investigates a fully distributed cooperation scheme for networked mobile manipulators. To achieve cooperative task allocation in a distributed way, an adaptation-based estimation law is established for each robotic agent to estimate the desired local trajectory. In addition, wrench synthesis is analyzed in detail to lay a solid foundation for tight cooperation tasks. Together with the estimated task, a set of distributed adaptive controllers is proposed to achieve motion synchronization of the mobile manipulator ensemble over a directed graph with a spanning tree irrespective of the kinematic and dynamic uncertainties in both the mobile manipulators and the tightly grasped object. The controlled synchronization alleviates the performance degradation caused by the estimation/tracking discrepancy during the transient phase. The proposed scheme requires no persistent excitation condition and avoids the use of noisy Cartesian-space velocities. Furthermore, it is independent from the object's center of mass by employing formation-based task allocation and a task-oriented strategy. These attractive attributes facilitate the practical application of the scheme. It is theoretically proven that convergence of the cooperative task tracking error is guaranteed. Simulation results validate the efficacy and demonstrate the expected performance of the proposed scheme.
	\end{abstract}
	
	
	\IEEEpeerreviewmaketitle
	
	%%%%%%%%%%%%%%%%%%%%%%%%%%%%%%%%%%%%%%%%%%%%%%%%%%%%%%%%%%%%%%%%%%%%%%%%%%%%%%%%
	\section{Introduction}
		\IEEEpubidadjcol
	\IEEEPARstart{M}{obile} manipulator, which combines the manipulation dexterity in robotic arm and the maneuverability in mobile platform, tends to be far more versatile than the conventional base-fixed counterpart due to its enlarged workspace and the potential for wider application scenarios, e.g, part transfer, rescue and remote maintenance in outdoor environment, etc. ~\cite{fruchard2006framework}. Based on that, multiple mobile manipulator ensemble has drawn increasing attention of the research community in recent years owing to its ability to perform more complex tasks such as transporting or assembling large and heavy objects that cannot be achieved by a single mobile robot. While these attractive advantages reveal a major increase of complexity for modelling and controlling such systems, especially for the case considered in this paper that a team of uncertain nonholonomic mobile manipulators cooperate to grasp and manipulate an unknown object. Introduction of the mobile platform, typically a nonholonomic vehicle, not only creates high degree of redundancy but also imposes nonintegrable constraints on the kinematics, which hinders direct control of the whole system and restricts its instantaneous motion capability. In addition, interactions between the mobile platform and the manipulator necessitate the integrated modelling method of both the system dynamics and kinematics. Furthermore, the tightly grasped object and the mobile manipulator ensemble form connected kinematics with star topology, which leads to the imposition of a set of kinematic/dynamic constraints on each mobile manipulator and the degradation of the degree of freedom. This will be accompanied by the generation of internal forces that needs careful regulation. Ignoring the control of these internal forces may result in grasp failure or unrecoverable damage to the end-effector (EE) or the object.
	
	\IEEEpubidadjcol
	
	The core problem in multi-robot manipulation besides the modelling complexity mentioned above lies in the establishment of a fully distributed scheme for the inherently centralized cooperation task, especially under certain communication constraint and ubiquitous model uncertainties. Note that the scheme presented in this paper may easily be extended to other cases by releasing some of the considered constraints.
	\subsection{Related Work}
    Cooperation and coordination of multi-agent system have been well recognized as an important technique to enhance flexibility and improve efficiency~\cite{dai2017distributed}. The endeavor to achieve manipulation and transportation of an object by multi-arm system in a cooperative manner generally comprises two control schemes: centralized control and distributed control~\cite{cao2013overview}. Under the centralized architecture, a global coordinator either in leader robot or in another host computer facilitates object-oriented control with the help of available global states of the robotic system. Force/position control~\cite{ren2017adaptive} and impedance control~\cite{caccavale2008six, ren2016biomimetic} are both extensively utilized to achieve safe interaction between the dual-arm manipulators and the grasped object. Extension of this cooperative manner to multiple mobile manipulators~\cite{li2010adaptive} and its model-based control can benefit from the comprehensive interaction dynamic model from the perspective of kinematic constraint~\cite{erhart2016model}. Although centralized architecture shows sufficient power to control multi-robot system and can easily incorporate single-robot control strategy, assumption of the existence of a central station makes the whole system more vulnerable and its malfunction will lead to the breakdown of the whole system~\cite{yan2013survey}. On the contrary, as a more promising and preferable alternative, distributed approach predominates when robot collectives are subjected to some inevitable physical constraints such as communication limitations. Specifically, it is unreasonable to assume there exists a coordinator for the case studied here since all team members are mobile.
    
    To achieve distributed control of a multi-arm system, leader-follower approach is an option generally associated with the schemes that are devoted to reducing the communication burden while trying to achieve as far as possible. Based on the diagram of impedance dynamics and leader-follower scheme, coordinated motion control for multiple mobile manipulators is employed to achieve cooperative object manipulation~\cite{kume2007coordinated}. Inspired by a team of people moving a table, the followers can implement similar impedance law as the leader's either by estimating the leader's desired motion~\cite{tsiamis2015cooperative} or by taking the contact force as the leader's motion intention~\cite{bai2010cooperative}. This innovation enables the whole system to work under implicit communication. Another interesting work~\cite{wang2016force} that does not require communication achieves force coordination between leader and follower only by measuring the object's motion as the feedback. However, the assumptions that all followers act passively in the transport task in~\cite{kume2007coordinated,tsiamis2015cooperative}, that desired velocity of the grasped object is constant and available to each agent in~\cite{bai2010cooperative}, that the attachment points of the collectives are centrosymmetric around the center of mass (COM) of the object in~\cite{wang2016force}, act as the primary factors that hind the applications of their works to our case.
    
    The idea that successful object transportation in real application is generally strongly related to the robot formation has also been continuously inspiring related works~\cite{eoh2011multi, alonso2017multi}. Along with the rapid advances in graph theory and control philosophy of the multi-agent systems, distributed scheme under explicit communication plays an important role in the formation control of multiple mobile robots~\cite{peng2013leader, montijano2016vision}. The challenge existing in formation-based transport task for multiple mobile manipulator ensemble lies in the design of a distributed control law to achieve a global behavior in cooperative manner with limited local information and constrained communication~\cite{antonelli2013interconnected}. A typical schema~\cite{marino2017distributed} employs a set of distributed controller/observer to achieve relative formation of the multi-arm system. Convergent estimation of the collective states by local observer bridges the gap between the local control and the global cooperative behavior, thus a totally distributed cooperation is achieved~\cite{antonelli2014decentralized}.
    
    To further maintain high performance when the mobile robot team executes tightly cooperative task whilst suffering from the inevitable uncertainties of the robot dynamics, adaptive mechanism is introduced into the architecture, based on either impedance control~\cite{li2013decentralised} or force/position control~\cite{marino2017distributed}. More complicate case in which the dynamic uncertainty and communication constraints (e.g. the jointly communication topology~\cite{dai2017distributed}) coexist can be easily tackled by embedding the parameter adaptation to the respective control scheme. Recent representative work~\cite{verginis2017robust} employs the robust adaptive control to concurrently address dynamic uncertainties and external disturbances. The dependence of the communication network and the costly force/torque sensor is further eliminated by introducing the assumption that all the robot agents know the desired trajectory and exact grasp parameter in advance. In addition, to cope with uncertainties of the grasped object, a distributed approach is presented in~\cite{franchi2015decentralized} to estimate the object's dynamic/kinematic parameters by properly moving the object or applying specific contact wrenches. Based on this estimation process, the cooperative manipulation of an unknown object can be further expected at the expense of a small bounded tracking error by a two-stage decentralized scheme~\cite{petitti2016decentralized, marino2018two}. While these works either assume that the object's COM is known to all robotic agents or employ a separate step to estimate the object's COM, the prerequisite persistent excitation condition may restrict their range of practical application.
    
    In addition, the robotic collectives in the above-mentioned works are free of kinematic uncertainties. However, cooperative transport task is very sensitive to kinematic uncertainties of the interconnected system. Since the manipulators rigidly contact with the object, small kinematic discrepancy may lead to large tracking error and build-up of the internal force. Adaptability to kinematic uncertainties endows the multiple robotic system with improved intelligence and flexibility. All of these demonstrate the necessity of handling the kinematic uncertainties with care.
    \subsection{Contribution}
    % The works devoted to the solution of both kinematic and dynamic uncertainties are not applicable to our case since all the robotic agents in these works have access to the desired trajectory and only cooperate loosely, i.e. only kinematic constraint is imposed on each individual motion. However, 
    Multi-arm manipulation is associated with tight cooperation in which both the dynamic and kinematic constraints are applied to each of the mobile manipulators. In addition, the cooperative task should be well allocated among the robotic agents also in a distributed way. In light of the above discussions, this study contributes a fully distributed control scheme for a team of networked nonholonomic mobile manipulators (NMMs) to cooperatively transport an unknown object with the following advantages which distinguish our proposed scheme from the existing approaches:

      \begin{itemize}
      \item [1)] 
      Uncertain dynamics/interconnected kinematics and limited communication are addressed comprehensively based on adaptive control.       
      \item [2)]
      Task allocation and cooperative control are fully distributed. Synchronization idea is fulfilled through the design of the whole control scheme, which can alleviate the performance degradation caused by the estimation and tracking discrepancy during transient phase.
      \item [3)]
      Persistent excitation condition and noisy Cartesian-space velocity are totally avoided.
      \item [4)]
      Independence from the coordinate attached to the object's COM by the task-oriented strategy and formation-based idea facilitates the practical implementation.
    \end{itemize}
    
		   		\begin{figure*}[ht]
		\sf
		\centering
		\normalsize
		\includegraphics[width=0.76\textwidth]{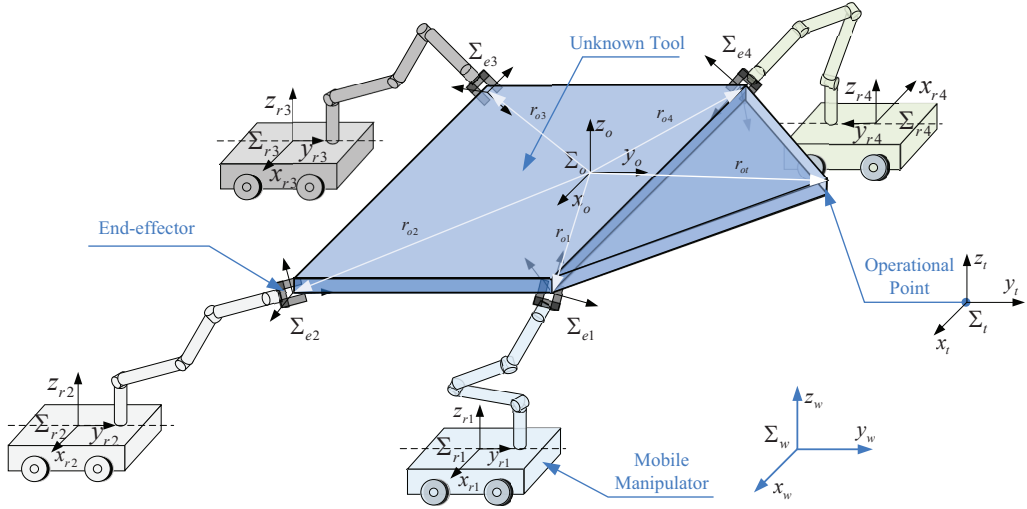}
		\caption{Networked mobile manipulators cooperatively transporting and manipulating an unknown tool \label{figure1}}
	\end{figure*}    
    
	\section{Problem Formulation}\label{sec:problemformulation}
	\subsection{Preliminaries} \label{sec:preliminaries}
	Consider $N$ mobile manipulators tightly grasping a common unknown object, as shown in Fig.~\ref{figure1}. Let ${\Sigma _{ei}}$ denote the frame fixed to the end-effector of the \emph{i}th mobile manipulator. Furthermore, the object frame ${\Sigma _o}$ and the cooperative task frame ${\Sigma _t}$ are two frames both attached to the object and their origin are chosen so as to coincide with the object's COM and the operational point respectively. All quantities are expressed with respect to the world frame ${\Sigma _w}$, unless otherwise stated. For each mobile manipulator, the mounted manipulator is considered as a holonomic system while the mobile platform is assumed to be nonholonomic. Throughout this paper, ${0_{m \times m}}$ represents $m \times m$ null matrix, ${I_m}$ denotes the $m \times m$ identity matrix and ${1_m} = {[1,...,1]^T} \in {\Re ^m}$ (${0_m} = {[0,...,0]^T} \in {\Re ^m}$) is $m \times 1$ column vector with all elements equal to 1 (0). The Cartesian-space variable ${x_{sub}} = {\left[ {p_{sub}^T,o_{sub}^T} \right]^T} \in {\Re ^m}$ can be split into translational part ${p_{sub}} \in {\Re ^3}$ and rotational part ${o_{sub}} \in {\Re ^3}$ in the case of ${m = 6}$.
 
	 \emph{Assumption 1}: The unknown object is rigid and the networked mobile manipulators grasp the object tightly so that there is no relative motion between the end-effectors and the object. Grasp strategy is not discussed here. For cooperative grasp strategy, the readers are referred to~\cite{basile2013decentralized, music2017robot}.\label{assumption1}
	 
    \emph{Assumption 2}: Each mobile manipulator has access to its end-effector position. This can be achieved by local on-board camera with localization.
	
	\subsection{Kinematics and dynamics of the interconnected system} \label{sec:KDIS}
	Denote by ${q_i} = {[q_{v,i}^T,q_{m,i}^T]^T} \in {\Re ^{{n_r}}}$ the generalized coordinates of the \emph{i}th mobile manipulator with ${q_{v,i}} \in {\Re ^{{n_v}}}$  representing the position and orientation of the mobile platform and  ${q_{m,i}} \in {\Re ^{{n_m}}}$ describing the joint angle vector of the mounted manipulator, and ${n_r} = {n_v} + {n_m}$ .
	
	The nonholonomic constraint acting on the mobile platform can be expressed as~\cite{peng2014robust}:
	\begin{align}
	\label{eq:1}
    {A_{v,i}}({q_{v,i}}){\dot q_{v,i}} = {0_{{n_c}}}
	\end{align}
	where ${A_{v,i}}({q_{v,i}}) \in {\Re ^{{n_c} \times {n_v}}}$  denotes the constraint matrix of the mobile platform. Constraint equation~\eqref{eq:1} implies that there exists a reduced vector ${\zeta _{v,i}} \in {\Re ^{{n_v} - {n_c}}}$ , such that
	\begin{align}
	\label{eq:2}
    {{\dot q_{v,i}} = {H_{v,i}}({q_{v,i}}){\dot \zeta _{v,i}}}
	\end{align}
	where ${H_{v,i}}({q_{v,i}})$ satisfies that ${H_{v,i}^T({q_{v,i}})A_{v,i}^T({q_{v,i}}) = {0_{{n_v} - {n_c}}}}$. Then a reduced vector ${\zeta _i} = {[\zeta _{v,i}^T,q_{m,i}^T]^T} \in {\Re ^{{n_r} - {n_c}}}$  can be defined, which will be used in the sequel.
	
	Let ${x_{e,i}} \in {\Re ^m}$ denote the EE pose vector of \emph{i}th mobile manipulator and it is related to the generalized coordinates by
	\begin{align}
	\label{eq:3}
    {{\dot x_{e,i}} = {J_i}({q_i}){\dot q_i} = {J_{e,i}}({\zeta _i}){\dot \zeta _i}}
	\end{align}
	where  ${{J_i}({q_i}) \in {\Re ^{m \times {n_r}}}}$ is the whole mobile manipulator Jacobian matrix. ${{J_{e,i}} \in {\Re ^{m \times ({n_r} - {n_c})}}}$  is the reduced Jacobian matrix that will be defined later. $m \le {n_r} - {n_c}$  and equality holds for the non-redundant mobile manipulator.
	
	\emph{Property 1}: The kinematics~\eqref{eq:3} linearly depends on a kinematic parameter vector ${{\theta _{k,i}} = {[{\theta _{k1,i}},...,{\theta _{k{p_{k,i}},i}}]^T} \in {\Re ^{{p_{k,i}}}}}$, such as joint offsets and link lengths of the manipulator~\cite{cheah2006adaptive}:
	\begin{align}
	\label{eq:4}
    {{\dot x_{e,i}} = {J_{e,i}}({\zeta _i},{\theta _{k,i}}){\dot \zeta _i} = {Y_{k,i}}({\zeta _i},{\dot \zeta _i}){\theta _{k,i}}}
	\end{align}
	where ${{Y_{k,i}}({\zeta _i},{\dot \zeta _i}) \in {\Re ^{m \times {p_{k,i}}}}}$ is the kinematic regressor matrix.
	
	 ${x_{obj}} \in {\Re ^m}$ is defined as the coordinate vector of the object's COM and it is assumed that ${\dot x_{obj}}$  is related to ${\dot x_{e,i}}$ by
	\begin{align}
	\label{eq:5}
    {{\dot x_{e,i}} = {J_{o,i}}{\dot x_{obj}}}
	\end{align}
	where ${{J_{o,i}} = [{I_3}, - S({r_{oi}});{0_3},{I_3}]}$ denotes the object Jacobian matrix. ${S(\cdot)}$ is the skew-symmetric operator and ${r_{oi}}$ represents the vector from object's COM to the corresponding contact point. Definitions of ${r_{oi}}$ for ${i=1,2,...,N}$ are presented in Fig.~\ref{figure1}. Here the object's COM is introduced to facilitate the force analysis and will be avoided in the controller design.
	
% 	Relation between  ${\dot x_{obj}}$ and the velocity vector of the object's operational point ${\dot x_t} \in {\Re ^m}$  can be expressed as
% 	\begin{align}
% 	\label{eq:6}
%     {{\dot x_{obj}} = R({x_t}){\dot x_t}}
%     \end{align}
%     where $R = [{I_3},S({r_{ot}});{0_3},{I_3}]$ is the invertible mapping matrix. Definitions of  ${r_{ot}}$  and  ${r_{oi}}$ are presented in Fig.~\ref{figure1}.
    
    Assumption 1 imposes the following kinematic constraints on the relative motion of the attached end-effectors.
	\begin{align}
	\label{eq:7}
    {{x_{e,i}} = {x_{e,j}} + {{{\cal T}}_{ji}}}
    \end{align}
    where ${{{{\cal T}}_{ji}} = {[{({R_{w,i}}{}^i{r_{ji}})^T},{}^w\phi _{ji}^T]^T}}$  with ${{R_{w,i}}{}^i{r_{ji}}}$  and  ${{}^w{\phi _{ji}}}$ respectively representing the relative displacement and orientation, ${{R_{w,i}}}$ is the rotation matrix of ${\Sigma _{ei}}$ with respect to ${\Sigma _w}$. ${{}^i{r_{ji}}}$ denotes the vector pointing from the \emph{j}th grasp point to the \emph{i}th grasp point expressed in the frame ${\Sigma _{ei}}$. The constraint between the \emph{i}th grasp point and operational point can be expressed in a similar way, i.e. ${{x_{ei}} = {x_t} + {{{\cal T}}_{ti}}}$ with ${{{{\cal T}}_{ti}} = {[{({R_{w,t}}{}^t{r_{ti}})^T},{}^w\phi _{ti}^T]^T}}$.  ${{}^t{r_{ti}}}$ is assumed to be known to the robotic agent since it is fixed after the object is grasped.
    
    Dynamics of the \emph{i}th mobile manipulator in joint space can be expressed as
    \begin{align}
	\label{eq:8}
    {{M_i}({q_i}){\ddot q_i} + {C_i}({q_i},{\dot q_i}){\dot q_i} + {G_i}({q_i}) = {B_i}({q_i}){\tau _i} - A_i^T{F_i}}
    \end{align}
    where ${{M_i}({q_i}) = [{M_{v,i}},{M_{vm,i}};{M_{mv,i}},{M_{m,i}}] \in {\Re ^{{n_r} \times {n_r}}}}$ denotes the symmetric positive definite inertial matrix and ${{C_i}({q_i},{\dot q_i}) = [{C_{v,i}},{C_{vm,i}};{C_{mv,i}},{C_{m,i}}] \in {\Re ^{{n_r} \times {n_r}}}}$ is the Coriolis and Centrifugal matrix. ${{M_{vm,i}}{\ddot q_{m,i}}}$ and ${{M_{mv,i}}{\ddot q_{v,i}}}$ represent the interaction inertia torques between manipulator and mobile platform, ${G_i}({q_i}) = {[G_{v,i}^T,G_{m,i}^T]^T} \in {\Re ^{{n_r}}}$ is the gravitational torque vector; ${{B_i}({q_i}) = diag[{B_{v,i}},{B_{m,i}}] \in {\Re ^{{n_r} \times p}}}$ is the input transformation matrix for the whole mobile manipulator; ${{\tau _i} = {[\tau _{v,i}^T,\tau _{m,i}^T]^T} \in {\Re ^p}}$ denotes the input torques; ${{F_i} = {[\lambda _{v,i}^T,F_{e,i}^T]^T} \in {\Re ^{{n_c} + m}}}$ in which ${\lambda _{v,i}} \in {\Re ^{{n_c}}}$ represents the Lagrange multiplier associated with the nonholonomic constraint and ${F_{e,i}} \in {\Re ^m}$ denotes the external wrenches exerted by the holonomic constraint. ${{A_i} = [{A_{v,i}},0;{J_{v,i}},{J_{m,i}}] \in {R^{({n_c} + m) \times {n_r}}}}$ in which ${{J_{v,i}}}$ and ${{J_{m,i}}}$ respectively represent the Jacobian matrices of the mobile base and the mounted manipulator with opportune dimensions.
    
    Considering~\eqref{eq:2} and its derivative and multiplying both sides of~\eqref{eq:8} by $\left[ {H_{v,i}^T,0;0,I} \right]$ yields the following reformulation:
    \begin{align}
	\label{eq:9}
    {{M_{r,i}}{\ddot \zeta _i} + {C_{r,i}}{\dot \zeta _i} + {G_{r,i}} = {B_{r,i}}{\tau _i} - J_{e,i}^T{F_{e,i}}}
    \end{align}    
    where ${{M_{r,i}}}$, ${{B_{r,i}}}$, ${{C_{r,i}}}$ and ${{G_{r,i}}}$ are given in Appendix A. Then the nonholonomic constraint force ${A_{v,i}^T{\lambda _{v,i}}}$ in~\eqref{eq:8} can be eliminated. ${{J_{e,i}} = [{J_{v,i}}{H_{v,i}},{J_{m,i}}] \in {\Re ^{m \times ({n_r} - {n_c})}}}$.
    
    Dynamics of the grasped object can be described by:
    \begin{align}
	\label{eq:10}
    {{M_o}({x_{obj}}){\ddot x_{obj}} + {C_o}({x_{obj}},{\dot x_{obj}}){\dot x_{obj}} + {g_o}({x_{obj}}) = {F_o}}
    \end{align}
    where ${{M_o}({x_{obj}}) \in {\Re ^{m \times m}}}$ denotes the inertial matrix and is assumed to be bounded and symmetric positive definite, ${{\lambda _{o\min }}{I_m} \le {M_o} \le {\lambda _{o\max }}{I_m}}$, where ${\lambda _{o\min }}$ and ${\lambda _{o\max }}$ represent the minimum and maximum eigenvalues of ${{M_o}}$;  ${{C_o}({x_{obj}},{\dot x_{obj}})}$ is the ${m \times m}$ Coriolis and Centrifugal matrix and ${{g_o}({x_{obj}}) \in {\Re ^{m \times m}}}$ represents the gravitational vector. ${F_o} \in {\Re ^m}$ is the resultant wrench acting on the object's COM by the multiple mobile manipulator ensemble.
    
    By virtual of kineto-statics duality, the resultant wrench ${F_o}$ acting on the object's COM satisfies the following relation:
    \begin{align}
	\label{eq:11}
    {{F_o} = {G_o}{F_e}}
    \end{align}
    where ${G_o} = J_o^T = [J_{o,1}^T,J_{o,2}^T,...,J_{o,N}^T] \in {\Re ^{m \times Nm}}$ is the well-known grasp matrix, ${F_e} = {[F_{e,1}^T,F_{e,2}^T,...,F_{e,N}^T]^T} \in {\Re ^{Nm}}$ is the collective vector consisting of all generalized wrenches exerted by the mobile manipulators on the object, which can be measured by the force/torque sensor mounted at the wrist of each manipulator. ${F_o}$ can also be expressed as ${F_o} = \sum\nolimits_{i = 1}^N {{F_{ce,i}}}$ where ${{F_{ce,i}} = J_{o,i}^T{F_{e,i}}}$ is the wrench indirectly applied on the object's COM by \emph{i}th end-effector and it can be decomposed into two orthogonal components, one is motion-induced wrench ${{F_{E,i}} \in {\Re ^m}}$ (MW) which contributes to the motion of the grasped object, the other one is the internal wrench ${{F_{I,i}} \in {\Re ^m}}$ (IW) which contributes to the build-up of the stress in the object. This relation can be expressed as: ${{F_{ce,i}} = {F_{E,i}} + {F_{I,i}}}$.
    
%     The property of the internal wrenches results in the following equation
%     \begin{align}
% 	\label{eq:14}
%     {\sum\nolimits_{i = 1}^N {{F_{I,i}}}  = 0}
%     \end{align}
    
    Since the internal wrenches cancel each other, i.e. ${\sum\nolimits_{i = 1}^N {{F_{I,i}}}  = 0}$, the object dynamics~\eqref{eq:10} can be rewritten as
    \begin{align}
	\label{eq:15}
    {{M_o}({x_{obj}}){\ddot x_{obj}} + {C_o}({x_{obj}},{\dot x_{obj}}){\dot x_{obj}} + {g_o}({x_{obj}}) = \sum\nolimits_{i = 1}^N {{F_{E,i}}}}
    \end{align}

    Suppose that a desired load distribution among the robot team is described by a set of positive definite diagonal matrices ${\beta _i}(t) \in {\Re ^{m \times m}}$ for $i = 1,2,...,N$ with the assumption that $\| {{{\dot \beta }_i}(t)} \| \le {d_i}$ (${d_i}$ is a positive constant)~\cite{liu1998decentralized}:
    \begin{align}
	\label{eq:16}
    {{F_{E,i}} = {\beta _i}(t)[{M_o}({x_{obj}}){\ddot x_{obj}} + {C_o}({x_{obj}},{\dot x_{obj}}){\dot x_{obj}} + {g_o}({x_{obj}})]}
    \end{align}
    
    An important physical property regarding these load distribution matrices is that ${\sum\nolimits_{i = 1}^N {{\beta _i}(t)}  = {I_m}}$. Then incorporating~\eqref{eq:16} and above-mentioned properties into~\eqref{eq:9} gives the complete dynamical equations of the interconnected system in a distributed fashion:
     \begin{align}
	\label{eq:19}
    {{M_{s,i}}{\ddot \zeta _i} + {C_{s,i}}{\dot \zeta _i} + {G_{s,i}} = {B_{r,i}}{\tau _i} - J_{\phi ,i}^T{F_{I,i}}}
    \end{align}   
    where 
    \begin{align*}
    \begin{split}
    &{{M_{s,i}} = {M_{r,i}} + {\beta _i}(t)J_{\phi ,i}^T{M_o}{J_{\phi ,i}}},\\
    &{{C_{s,i}} = {C_{r,i}} + {\beta _i}(t)[J_{\phi ,i}^T{C_o}{J_{\phi ,i}} + J_{\phi ,i}^T{M_o}\frac{{d{J_{\phi ,i}}}}{{dt}}]},\\
    &{{G_{s,i}} = {G_{r,i}} + {\beta _i}(t)J_{\phi ,i}^T{g_o}},\quad{J_{\phi ,i}} = J_{o,i}^\dag {J_{e,i}}.
    \end{split}
    \end{align*}
    \emph{Property 2}: ${{\dot M_{s,i}} - 2{C_{s,i}} - {\dot \beta _i}J_{\phi ,i}^T{M_o}{J_{\phi ,i}}}$ is a skew symmetric matrix so that ${{\upsilon ^T}[{\dot M_{s,i}} - 2{C_{s,i}} - {\dot \beta _i}J_{\phi ,i}^T{M_o}{J_{\phi ,i}}]\upsilon  = 0}$ for all $\upsilon  \in {\Re ^{({n_r} - {n_c})}}$. Please see Appendix A for the proof.
    
    Then the following inequality holds since the robot work in finite joint space
    \begin{align}
	\label{eq:20}
    {\left\| {{{\dot \beta }_i}J_{\phi ,i}^T{M_o}{J_{\phi ,i}}} \right\| \le {d_i}{\lambda _{o\max }}\left\| {J_{\phi ,i}^T{J_{\phi ,i}}} \right\| = {\vartheta _i}}
    \end{align}
    where ${\vartheta _i}$ is a positive constant that denotes the upper bounds of ${\| {{{\dot \beta }_i}J_{\phi ,i}^T{M_o}{J_{\phi ,i}}} \|}$.
    
    \emph{Property 3}: The nonlinear dynamics linearly depends on a dynamic parameter vector ${{\theta _{d,i}} = {[{\theta _{d1,i}},...,{\theta _{d{p_{d,i}},i}}]^T} \in {\Re ^{{p_{d,i}}}}}$ which is composed of physical parameters of the object and mobile manipulator, which gives rise to
    \begin{align}
	\label{eq:21}
    {{M_{s,i}}{\ddot \zeta _i} + {C_{s,i}}{\dot \zeta _i} + {G_{s,i}} = {Y_{d,i}}({\zeta _i},{\dot \zeta _i},{\ddot \zeta _i},{\beta _i}){\theta _{d,i}}}
    \end{align}
    where ${{Y_{d,i}}({\zeta _i},{\dot \zeta _i},{\ddot \zeta _i},{\beta _i}) \in {\Re ^{({n_r} - {n_c}) \times {p_{d,i}}}}}$ is the dynamic regressor matrix.
    
\begin{remark}
To avoid input transformation uncertainty which is associated with kinematic uncertainty of the mobile base, ${{\zeta _i}}$ can be defined to coincide with the actuation space of the mobile manipulator, i.e. ${{\zeta_i} ={[{q_{R,i}},{q_{L,i}},{q_{m1,i}},...{q_{m{n_r}i}}]^T}}$ for a two-wheel mobile platform where ${{q_{R,i}}}$ and ${{q_{L,i}}}$ are the two independent driving wheels. Thus, ${{B_{r,i}} = {I_{({n_r} - {n_c})}}}$ and the uncertainties in the input transformation matrix are eliminated.
\end{remark}
	
	\subsection{Communication topology} \label{sec:CT}	
	As is commonly done in the distributed control, a graph ${{\cal G}} = ({{\cal V}},{{\cal E}})$ is employed here to describe the communication topology among the $N$ mobile manipulator systems where the vertex set ${{\cal V}} = \{ 1,2,...,N\}$ represents all the robots in the network and the edge set ${{\cal E}} \subseteq {{\cal V}} \times {{\cal V}}$ denotes the communication interaction between the robots. The edge $(i,j)$ in directed graph indicates that robot $i$ can receive information from robot $j$ but not vice versa. Neighbors of robot $i$ form a set ${{{\cal N}}_i} = \{ j|(i,j) \in {{\cal E}},j \in {{\cal V}}\} $. The $N \times N$ adjacency matrix ${{\cal A}} = [{a_{ij}}]$ associated with this graph is defined as ${a_{ij}} = 1$ if $j \in {{{\cal N}}_i}$ and ${a_{ij}} = 0$ otherwise. Additionally, self-loops are not contained, i.e. ${a_{ii}} = 0$. Then the Laplacian matrix ${{\cal L}} = [{l_{ij}}] \in {\Re ^{N \times N}}$ can be defined as
	\begin{align}
	\label{eq:22}
    {{l_{ij}} = \left\{ \begin{array}{l}
\sum\nolimits_{k = 1}^N {{a_{ik}}} \qquad {\textrm{if }}i = j\\
 - {a_{ij}} \qquad \qquad {\textrm{otherwise}}
\end{array} \right.}
    \end{align}
    
    Several properties associated with the Laplacian matrix are listed in the following lemma.
	
 	\emph{Lemma 1}: The matrix $\left( {{{\cal L}} + {{\cal B}}} \right)$ is nonsingular and ${{\cal Q}} = {{\cal P}}\left( {{{\cal L}} + {{\cal B}}} \right) + {\left( {{{\cal L}} + {{\cal B}}} \right)^T}{{\cal P}}$ is positive definite with following definitions: ${{\cal P}} = diag\left\{ {{{{\cal P}}_i}} \right\} = diag\left\{ {{1 \mathord{\left/
  {\vphantom {1 {{w_i}}}} \right.
  \kern-\nulldelimiterspace}{{w_i}}}}\right\}$ and ${{{\cal B}} = diag\left\{{{b_i}}\right\}}$ where $w={\left[{{w_1},...,{w_N}} \right]^T} = {\left( {{{\cal L}}+{{\cal B}}}\right)^{- 1}}{1_N}$ and ${b_i}=1$ is employed to indicate that the ${x_{t,d}}$ is accessible to the \emph{i}th (${i_l}$) robotic agent, otherwise ${b_i} = 0$~\cite{qu2009cooperative}.
      
    \emph{Assumption 3}: In this paper, the graph that the $N$ mobile manipulators interact on is directed with a spanning tree whose root node has direct access to the desired trajectory of the operational point, i.e. ${x_{td}}$.
    
   This assumption implies that only a small subset of the robotic agents has direct access to the desired cooperative trajectory ${x_{td}}$. It is a more general case with less restriction in real application scenarios, yet, may complicate the controller design since the agents that cannot obtain ${x_{td}}$ should estimate it to achieve cooperative task by utilizing only locally available signals.
 
 Based on the models, assumptions and interaction topology discussed above, the control objective is to design a set of distributed controllers ${\tau _i}$ for the multiple mobile manipulator system to cooperatively transport an object irrespective of uncertain dynamics and closed-chain kinematics such that:
      \begin{itemize}
      \item [1)] 
      Operational point on the manipulated object could track a desired designated trajectory.       
      \item [2)]
      Estimation and motion tracking of the networked mobile manipulators are synchronized. 
      \item [3)]
      All the signals in the closed-loop interconnected system remain bounded.
    \end{itemize}
    
		\begin{figure*}[ht]
		\sf
		\centering
		\normalsize
		\includegraphics[width=0.8\textwidth]{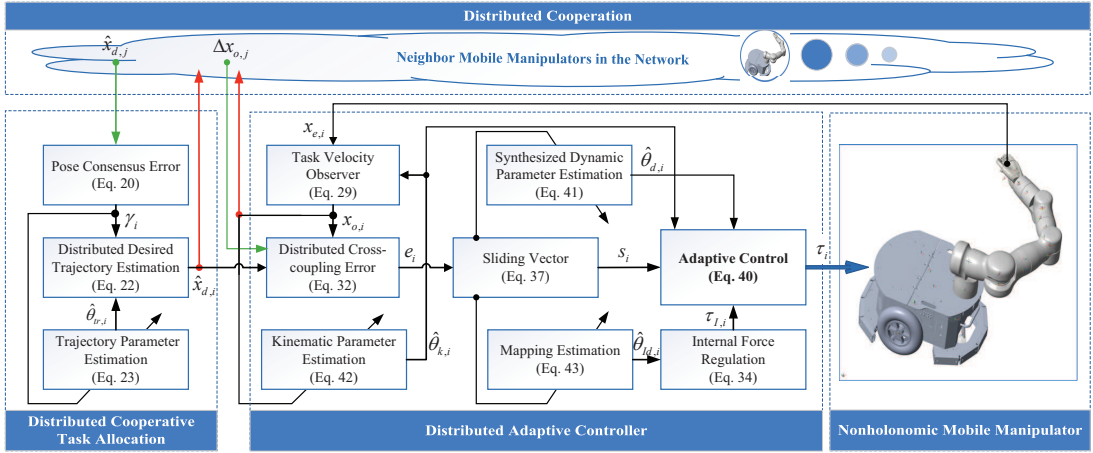}
		\caption{Fully distributed cooperation scheme for the networked mobile manipulator\label{figure2}}
	\end{figure*}
	
	\section{Distributed Adaptive Cooperation}\label{sec:dac}
	This section is devoted to the formulation of the fully distributed cooperation scheme, whose main structure is shown in Fig.~\ref{figure2}. To enable the networked robotic system to perform a tight cooperation task, i.e. object transportation discussed in this paper, the cooperative task should be well allocated to every agent in a distributed way. In addition, internal wrench emerging from constraint dynamics and the motion tracking control of each agent in the presence of kinematic/dynamic uncertainties both need careful treatment. Discussions of these three topics constitute the proposed scheme and are respectively presented in the following subsections.
	
	\subsection{Distributed cooperative task allocation} \label{sec:dcta}
	To achieve motion tracking control of the operational point, the task allocation for the end-effector of each mobile manipulator under limited communication can be interpreted as a rigid formation control problem, i.e. the operational point can be treated as a virtual leader that generates a desired trajectory and virtual followers are associated with the mobile manipulators that should be controlled to maintain the rigid formation in consideration of the rigid grasp.
	
    A continuous function usually can be approximated by a linear combination of a set of prescribed basis functions~\cite{yu2012adaptive}, i.e. ${f(t) = \sum\nolimits_{k = 1}^l {{f_{tr,k}}\left( t \right){\Theta _{tr,k}}}}$ . Then the desired trajectory for the operational point of the grasped object ${{x_{td}} = {[p_{td}^T,o_{td}^T]^T}}$ can be represented as
\begin{align}
	\label{eq:23}
	\begin{split}
    % {\begin{array}{c}
{x_{td}} =& {\left[ {{{{\cal F}}_{tr}}\left( t \right){{\bar \Theta }_{tr,1}},...,{{{\cal F}}_{tr}}\left( t \right){{\bar \Theta }_{tr,m}}} \right]^T}\\
 = &\left[ {\begin{array}{*{20}{c}}
{\left( {{I_3} \otimes {{{\cal F}}_{tr}}\left( t \right)} \right){\theta _{tr,p}}}\\
{\left( {{I_3} \otimes {{{\cal F}}_{tr}}\left( t \right)} \right){\theta _{tr,o}}}
\end{array}} \right] = \left( {{I_6} \otimes {{{\cal F}}_{tr}}\left( t \right)} \right){\theta _{tr}}
% \end{array}}
\end{split}
\end{align}
    where ${{{{\cal F}}_{tr}}\left( t \right){\rm{ = [}}{f_{tr,1}},{f_{tr,2}},...,{f_{tr,l}}{\rm{]}} \in {\Re ^l}}$ is the collective basis function that is assumed to be known to all robotic agents. ${{\bar \Theta _{tr,i}} = {[{\Theta _{tr,1}},...,{\Theta _{tr,l}}]^T} \in {\Re ^l}}$ is the constant parameter vector for the \emph{i}th component of ${{x_{td}}}$. ${{\theta _{tr}} = {[\theta _{tr,p}^T,\theta _{tr,o}^T]^T} = {[\bar \Theta _{tr,1}^T,...,\bar \Theta _{tr,m}^T]^T} \in {\Re ^{lm}}}$ denotes the constant parameters which is only available to the robotic agents marked by ${b_{{i_l}}} = 1$. This statement coincides with Assumption 3 and the definition of the matrix ${{\cal B}}$.
    
    \begin{remark}
It is worth mentioning here that employing~\eqref{eq:23} is reasonable to approximate the desired trajectories, especially in the context of the robot trajectory planning since they are typically planned as an interpolating polynomial function to guarantee the continuity of the velocity and acceleration.
\end{remark}

To facilitate the distributed estimation of the desired trajectory, the following two error variables are defined:
    \begin{align}
    \begin{split}
	\label{eq:24}
%    {\begin{array}{c}
{\varepsilon _{d,i}} =  &- ( {1 - {b_i}}) \left[ {( {{I_6} \otimes {{{\cal F}}_{tr}}(t)} ){{\hat \theta }_{tr,i}} + {{{\hat {\cal T}}}_{ti}}({{\hat o}_{td,i}})} \right]\\
 &+ {{\hat x}_{d,i}} - {b_i}( {{x_{td}} + {{{\cal T}}_{ti}}})
%\end{array}}
    \end{split}
    \end{align}
    \begin{align}
	\label{eq:25}
    {{\delta _i} = {\hat x_{d,i}} - {{{\cal T}}_{ti}} - {x_{td}}}
    \end{align}
  where ${{\varepsilon _{d,i}} = {[\varepsilon _{pd,i}^T,\varepsilon _{od,i}^T]^T}}$ is intermediate error variable. ${\delta _i} = {[\delta _{p,i}^T,\delta _{o,i}^T]^T}$ denotes the actual estimation error between each mobile manipulator and the desired trajectory. ${\hat x_{d,i}} = {[\hat p_{d,i}^T,\hat o_{d,i}^T]^T}$ represents the distributed estimation of the desired trajectory allocated to the end-effector of ith mobile manipulator, which together achieve ideal cooperative trajectory under rigid interconnection condition; ${{\hat \theta _{tr,i}} = {[\hat \theta _{tr,p,i}^T,\hat \theta _{tr,o,i}^T]^T}}$ and ${{\hat o_{td,i}}}$ are the estimates of the trajectory parameter vector ${{\theta _{tr}}}$ and ${{o_{td}}}$ by \emph{i}th mobile manipulator. ${{{\hat {\cal T}}_{ti}} = {[{({\hat R_{w,t}}({\hat o_{td,i}}){}^t{r_{ti}})^T},{}^w\phi _{ti}^T]^T}}$ is the estimate of ${{{{\cal T}}_{ti}}}$.   
  
  To achieve our control objective, the definition of standard local neighborhood consensus error~\cite{das2010distributed} is redesigned as:
      \begin{align}
	\label{eq:26}
	\begin{split}
    % {\begin{array}{c}
{\gamma _i} = &\sum\limits_{j \in {{{\cal N}}_i}} {\left[{{{\hat x}_{d,i}} - {{\hat x}_{d,j}} - {b_i}{{{\cal T}}_{ji}} - (1 - {b_i}){{{\hat {\cal T}}}_{ji}}} \right]} \\
 &+ {b_i}({{\hat x}_{d,i}} - {{{\cal T}}_{ti}} - {x_{dt}})
% \end{array}}
\end{split}
    \end{align}
    where ${{{\hat {\cal T}}_{ji}} = {[{({\hat R_{w,i}}({\hat o_{d,i}}){}^i{r_{ji}})^T},{}^w\phi _{ji}^T]^T}}$ is the estimate of ${{{{\cal T}}_{ji}}}$. ${{\gamma _i} = {[\gamma _{p,i}^T,\gamma _{o,i}^T]^T}}$ can be split into ${\gamma _{p,i}} \in {\Re ^3}$ and ${\gamma _{o,i}} \in {\Re ^3}$.
    
    Together with~\eqref{eq:25}, ~\eqref{eq:26} can be reformulated as:
     \begin{align}
 	\label{eq:27}
     {{\gamma _i} = \sum\limits_{j \in {{{\cal N}}_i}} {\left[ {{\delta _i} - {\delta _j} - (1 - {b_i}){{{\tilde {\cal T}}}_{ji}}} \right]}  + {b_i}{\delta _i}}
    \end{align}
    where ${{{\tilde {\cal T}}_{ji}} = {{\hat {\cal T}}_{ji}} - {{{\cal T}}_{ji}} = {[{(({\hat R_{w,i}} -{R_{w,i}}){}^i{r_{ji}})^T},0_3^T]^T}}$ is the estimation error of ${{\tilde {\cal T}}_{ji}}$.
    
    Then, the distributed estimation law of the local desired trajectory and its parameter update law are proposed as:
 \begin{align}
 	\label{eq:28}
 	\begin{split}
% {\begin{array}{c}
{{\dot{\hat x}}_{di}} = & - \kappa {{{\cal P}}_i}{\gamma _i} + {b_i}\left[ {( {{I_6} \otimes {{{\dot {\cal F}}}_{tr}}} ){\theta _{tr}} - {{{\dot {\cal T}}}_{ti}}} \right]\\
&+\left( {1 - {b_i}} \right)\left[ {( {{I_6} \otimes {{{\cal F}}_{tr}}} ){{\dot{\hat \theta}}_{tr,i}} + ( {{I_6} \otimes {{{\dot {\cal F}}}_{tr}}} ){{\hat \theta }_{tr,i}}} \right]\\
&+\left( {1 - {b_i}} \right){{{\dot{\hat {\cal T}}}}_{ti}}({{\hat o}_{td,i}},{{\hat \omega }_{td,i}})
% \end{array}}
\end{split}
\end{align}
\begin{align}
\label{eq:29}
{{\dot{\hat \theta}_{tr,i}} =  - {\Gamma _{tr,i}}{\left( {{I_6} \otimes {{{\cal F}}_{tr}}\left( t \right)} \right)^T}{\gamma _i}\qquad{\textrm{for }}{b_i} = 0}
\end{align}
where ${\kappa }$ and ${\Gamma _{tr,i}}$ the positive constant. ${{{\dot {\cal T}}_{ti}}}$ is the derivative of ${{{{\cal T}}_{ti}}}$ and it equals to ${{[{(S({\omega _{td}}){R_{w,t}}{}^t{r_{ti}})^T},0_3^T]^T}}$. ${\omega _{td}}$ denotes the first derivative of ${o _{td}}$ and this nomenclature applies to the definition of ${{\omega }_{td,i}}$ and ${\hat \omega }_{td,i}$.

 \begin{remark}
Since ${{{{\cal T}}_{ji}} = {[{({R_{w,i}}({o_{d,i}}){}^i{r_{ji}})^T},{}^w\phi _{ji}^T]^T}}$ is associated with ${{o_{di}}}$, thus is also not applicable to the robotic agents with ${b_i} = 0$. It should be noted that the pose (including the position and orientation) formation tracking control is achieved here, which distinguishes the above estimation law from the standard consensus error defined in~\cite{das2010distributed}.
\end{remark}

\begin{remark}
 Unlike other decentralized or distributed schemes \cite{li2013decentralised, liu1998decentralized} in which the desired cooperative trajectory for each robotic agent is assumed to be known to all, only locally available signals are utilized in the above estimation/update law~\eqref{eq:27}-~\eqref{eq:29}, thus endowing the whole mobile manipulator ensemble with the ability to work in a fully distributed way. Furthermore, the proposed estimation law does not require any persistent excitation condition of the desired trajectory and it is independent from the frame of the object's COM. These two wonderful properties that are also the objectives pursued through the following adaptive control design facilitate the whole scheme's practical implementation.
 \end{remark}
 \subsection{Internal wrench regulation}
  Before formulating the distributed internal wrench regulation, three constraints that a physically plausible internal wrench must obey are presented first~\cite{donner2018physically}:
\begin{align}
\label{eq:30}
    {\begin{array}{l}
{\left\| {{{\bar f}_{I,i}}} \right\|^2} \le \bar f_{s,i}^T{{\bar f}_{I,i}}\\
{\left\| {{{\bar \tau }_{fI,i}}} \right\|^2} \le \bar \tau _{sf,i}^T{{\bar \tau }_{fI,i}}\\
{\left\| {{{\bar \tau }_{I,i}}} \right\|^2} \le \bar \tau _{s,i}^T{{\bar \tau }_{I,i}}
\end{array}}
\end{align}
where ${{\bar F_{I,i}} = {[\bar f_{I,i}^T,\bar \tau _{I,i}^T]^T}}$ is the internal wrench mapped to the contact point. ${{\bar F_{s,i}} = {[\bar f_{s,i}^T,\bar \tau _{s,i}^T]^T}}$ denotes the wrench sensed by the force/torque sensor mounted on the \emph{i}th manipulator and it is equivalent to the interaction wrench ${F_{e,i}}$. ${\bar \tau _{sf,i}}$ is the torque induced by ${\bar f_{s,i}}$ and ${{\bar \tau _{fI,i}}}$ denotes component internal torque induced by ${\bar f_{s,i}}$.

Extracting internal wrench component from the interaction wrench (Wrench Decomposition) needs full contact wrench information and thus is inherently centralized. However, definition of the internal wrench is still a controversial and active research topic in multi-arm cooperation. Three typical wrench decomposition approaches, i.e. nonsqueezing specific pseudoinverse~\cite{walker1991analysis}, generalized Moore-Penrose inverse~\cite{li2010adaptive} and parametrized pseudoinverse~\cite{erhart2015internal, sieber2018human}, have been widely used in the internal wrench control. The above physical constraints are further taken into consideration to limit the range of the internal wrench~\cite{donner2018physically}.

Although it is in general not possible to achieve full decomposition of interactions into IWs leading to wrenches compensating each other and MWs contributing to the resulting wrench only without compensating parts~\cite{schmidts2016new}, the idea of a non-squeezing pseudoinverse~\cite{walker1991analysis} can be employed in wrench synthesis problem. Different from the statement in~\cite{walker1991analysis} that the non-squeezing load distribution is unique, we will show a more general formulation with non-squeezing effect in desired MWs. From the view of physical plausibility, the desired internal wrench cannot be arbitrary assigned once the desired load distribution is given. To avoid virtual wrenches that may appear in most of the internal wrench regulation schemes, the wrench decomposition constraints in wrench analysis~\cite{donner2018physically, schmidts2016new} are further incorporated in our proposed wrench synthesis.

To achieve desired grasp, wrench synthesis is discussed here and every solution mentioned above for the wrench decomposition can be straightforwardly applied. For optimized load distribution, readers are referred to~\cite{bais2015dynamic}. Here we adopt the nonsqueezing pseudoinverse~\cite{walker1991analysis} to formulate the wrench synthesis scheme. The desired MW can be given as
     \begin{align}
 	\label{eq:31}
     {{\bar F_{Ed,i}} = \left[ {\begin{array}{*{20}{c}}
{{\beta _i}{I_3}}&{{0_3}}\\
{ - {\beta _i}S({r_i})}&{{\beta _i}{I_3}}
\end{array}} \right]{F_{od}}}
    \end{align}
    where ${{F_{od}} = {[f_{od}^T,\tau _{od}^T]^T}}$ is the desired net wrench that should be applied to the object's COM. ${{\bar F_{Ed,i}}}$ denotes the desired MW mapped to the contact point. As to wrench synthesis, there is no need to keep consistency of the pseudoinverse matrix utilized in MW and IW, which leads to a more general synthesis formulation.

In this way, the designation of the desired MW and IW for each mobile manipulator is totally decoupled, which contributes to the reduction in the number of constraint equations and the ease of implementation. The constraints imposed on desired load distribution and desired internal wrench are reformulated as the following inequalities in terms of the wrench synthesis.
     \begin{align}
 	\label{eq:32}
     {\left\{ \begin{array}{l}
{\beta _i}{f_{od}}{f_{Id,i}} \ge 0\\
\left( {S({r_i}){\beta _i}{f_{od}} - {\beta _i}{\tau _{od}}} \right)\left( {S({r_i}){f_{Id,i}} - {\tau _{Id,i}}} \right) \ge 0
\end{array} \right.}
    \end{align}
    where ${{F_{Id,i}} = {[f_{Id,i}^T,\tau _{Id,i}^T]^T}}$ represents the desired internal wrench.

The following linearization is introduced to facilitate the internal wrench regulation:
 \begin{align}
 	\label{eq:33}
     {J_{\phi ,i}^T{F_{Id,i}} = {Y_{Id,i}}({\zeta _i},{\dot \zeta _i},{F_{Id,i}}){\theta _{Id,i}}}
    \end{align}
where ${{Y_{Id,i}}({\zeta _i},{\dot \zeta _i},{F_{Id,i}}) \in {\Re ^{({n_r} - {n_c}) \times {p_{Id,i}}}}}$ is the regressor matrix associated with the desired internal wrench and ${{\theta _{Id,i}} = {[{\theta _{Id,1,i}},{\theta _{Id,2,i}},...,{\theta _{Id,{p_{Id,i}},i}}]^T} \in {\Re ^{{p_{Id,i}}}}}$  is the linearized parameters.

The internal wrench regulation law ${{\tau _{I,i}}}$, that will be used in the sequel, can be given as:
 \begin{align}
 	\label{eq:34}
     {{\tau _{I,i}} = \hat J_{\phi ,i}^T{F_{Id,i}} + {\tau _{r,i}}}
    \end{align}
    where ${\hat J_{\phi ,i}^T}$ is the estimate of the Jacobian matrix transpose ${J_{\phi ,i}^T}$, ${{\tau _{r,i}} =  - {k_{r,i}}J_{e,i}^T{J_{e,i}}{s_i}}$ is the robust term and ${{k_{r,i}}}$ is a dynamically regulated positive gain whose range will be given in the sequel.
    \begin{remark}
    With above regulation law, the internal wrench tracking error only approaches to a neighborhood of the zero point if the P.E. condition is not satisfied. This is acceptable in most applications, especially for the multi-arm grasp, since internal wrench is only required to be regulated around a constant value to either avoid excessive stress or to provide sufficient grasp force~\cite{namvar2005adaptive}.
    \end{remark}
    
    \begin{remark}
    The decentralized schemes in the related literature that cope with the internal wrench regulation can be classified into two categories. In the first one~\cite{liu1998decentralized, lian2002semi}, the robotic agents cannot communicate with each other but are assumed to have access to the global force information, so these schemes can only be termed as semi-decentralized scheme. The other approach~\cite{marino2017distributed} uses the moment-based observer to estimate the net wrench acting on the grasped object exerted by the mobile manipulator ensemble. However, this is not applicable to our case because it requires the accurate dynamics and kinematics of the grasped object.  Different from the above-mentioned works in which internal force is calculated or estimated in a centralized way and an extra integral feedback of the internal force is applied to reduce the overshoot and limit the upper bound of the interaction force, we adopt the synchronization mechanism and small control gains instead. To maintain high tracking performance under multiple uncertainties even with small gains, adaptability studied in the next subsection is effective. This compromise endows the scheme with an attractive attribute that calculation of the internal force is avoided, thus maintaining the distributed manner of the whole scheme. For tight connection case considered in this paper, the internal wrench should be bounded as small as possible to avoid potential damage to the whole system, i.e. ${{F_{Id,i}} = 0}$.
    \end{remark}

    \subsection{Distributed adaptive control}
    In this section, a distributed adaptive control is presented to achieve allocated task tracking and task motion synchronization of the interconnected multiple mobile manipulator under uncertain close-chain kinematics and dynamics. This can be interpreted as ${\Delta {x_{e,i}},\Delta {\dot x_{e,i}} \to 0}$ and ${\Delta {x_{e,i}} - \Delta {x_{e,j}},\Delta {\dot x_{e,i}} - \Delta {\dot x_{e,j}} \to 0}$ after ${{\delta _i} \to 0}$, where ${\Delta {x_{e,i}} = {x_{e,i}} - {\hat x_{d,i}}}$ denotes the Cartesian-space tracking error of the \emph{i}th mobile manipulator. Different from state-of-art works~\cite{marino2017distributed, marino2018two}, motion synchronization should be achieved here to alleviate the transient performance degradation since multiple uncertainties exist in our case and the mobile manipulator ensemble are tightly interconnected through the object.
    
    To avoid the noisy Cartesian-space velocities, a distributed observer is presented as follows~\cite{liu2006adaptive}:
     \begin{align}
 	\label{eq:35}
     {{\dot x_{o,i}} = {\hat{\dot x}_{e,i}} - {\alpha _i}({x_{o,i}} - {x_{e,i}})}
    \end{align}
    where ${{\dot x_{o,i}} \in {\Re ^m}}$ denotes the observed Cartesian -space velocity, ${{\alpha _i}}$ is a positive design constant. ${{\hat{\dot x}_{e,i}} \in {\Re ^m}}$ is the estimated Cartesian-space velocity in the presence of kinematic uncertainties and it can be expressed by
         \begin{align}
 	\label{eq:36}
     {{\hat{\dot x}_{e,i}} = {\hat J_{e,i}}({\zeta _i},{\hat \theta _{k,i}}){\dot \zeta _i} = {Y_{k,i}}({\zeta _i},{\dot \zeta _i}){\hat \theta _{k,i}}}
    \end{align}
    where ${{\hat \theta _{k,i}}}$ is the estimated kinematic parameter vector and ${{\hat J_{e,i}}({\zeta _i},{\hat \theta _{k,i}})}$ is the estimated Jacobian matrix.
    
    Substituting~\eqref{eq:36} into~\eqref{eq:35} and using~\eqref{eq:3} yields the following closed-loop dynamics of the observer:
    \begin{align}
 	\label{eq:37}
     {{\dot{\tilde x}_{o,i}} =  - {\alpha _i}{\tilde x_{o,i}} + {Y_{k,i}}({\zeta _i},{\dot \zeta _i}){\tilde \theta _{k,i}}}
    \end{align}
    where ${{\tilde x_{o,i}} = {x_{o,i}} - {x_{e,i}}}$ denotes the observer error. ${{\tilde \theta _{k,i}} = {\hat \theta _{k,i}} - {\theta _{k,i}}}$ is the estimation error of the linearized kinematic parameters.
    
    To achieve both task tracking objective of each robot and synchronization objective among the robots, we first define a novel cross-coupling error ${{e_i} \in {\Re ^m}}$
    \begin{align}
 	\label{eq:38}
     {{e_i} = \Delta {x_{o,i}} + \sum\limits_{j \in {{{\cal N}}_i}} {\int_0^t {{\varepsilon _i}(\Delta {x_{o,i}} - \Delta {x_{o,j}})} }}
    \end{align}
    where ${\Delta {x_{o,i}} = {x_{o,i}} - {\hat x_{d,i}} \in {\Re ^m}}$, ${{\varepsilon _i}}$ is a positive constant.
    \begin{remark}
    Inspired by the centralized cross-coupling error proposed in~\cite{sun2002adaptive},~\eqref{eq:38} further takes the communication constraints into consideration and employs the observer error instead of tracking error to achieve distributed control and to deal with the kinematic uncertainties.
    \end{remark}
        \begin{remark}
  Motion synchronization of networked robotic agents is achieved here through explicit convergence of the distributed coupling error, which finds a balance between the synchronization behavior and the separation property suggested in~\cite{wang2013task, wang2017adaptive}.
    \end{remark}
    Then, define a Cartesian-space sliding variable ${S_{x,i}} \in {\Re ^m}$
    \begin{align}
 	\label{eq:39}
     {{S_{x,i}} = {\dot e_i} + {\Lambda _i}{e_i}}
    \end{align}
    where ${\dot e_i}$ is the time derivative of cross-coupling error ${{e_i}}$ and ${\Lambda _i}$ is the adjustable positive diagonal matrix.
    
    Considering the control objectives, reference joint velocity ${{\dot \zeta _{r,i}} \in {\Re ^{({n_r} - {n_c})}}}$ for the \emph{i}th mobile manipulator is defined as
    \begin{align}
 	\label{eq:40}
 	\begin{split}
    %  {\begin{array}{c}
{{\dot \zeta }_{r,i}} = &\hat J_{e,i}^\dag ({\zeta _i},{{\hat \theta }_{k,i}})\overbrace {[{{\dot{\hat x}}_{d,i}} - \sum\limits_{j \in {{{\cal N}}_i}} {{\varepsilon _i}(\Delta {x_{o,i}} - \Delta {x_{o,j}})}  - {\Lambda _i}{e_i}]}^{{{\dot x}_{pr,i}}}\\
 &+ \underbrace {{{\hat N}_{e,i}}{{({{\hat J}_{s,i}}{{\hat N}_{e,i}})}^\dag }({{\dot x}_{sr,i}} - {{\hat J}_{s,i}}{{\hat J}_{e,i}}^\dag {{\dot x}_{pr,i}})}_{{{\dot \zeta }_{sr,i}}}
% \end{array}}
\end{split}
    \end{align}
  where ${{\hat N_{e,i}}({\zeta _i},{\hat \theta _{k,i}}) = {I_{({n_r} - {n_c})}} - \hat J_{e,i}^\dag {\hat J_{e,i}}}$ denotes the null-space projector of the estimated Jacobian matrix ${{\hat J_{e,i}}}$ and ${\hat J_{e,i}^\dag ({\zeta _i},{\hat \theta _{k,i}}) \in {\Re ^{({n_r} - {n_c}) \times m}}}$ the pseudoinverse. ${{\dot x_{sr,i}}}$ denotes desired the local subtask of the \emph{i}th mobile manipulator which is implemented by utilizing the robotic agent's redundancy and ${{\hat J_{s,i}}}$ is the estimated subtask Jacobian. The above definition employs the multi-priority framework to maintain the designated task priorities under kinematic uncertainties.

%   To entitle the robotic system with the concurrent adaptability to both kinematic and dynamic uncertainties, various globally convergent adaptive controllers, generally classified into passivity-based control [30, 31] and inverse dynamic control [32], have been proposed by virtue of adaptive Jacobian technique and Cartesian-space position feedback.[34] presented an adaptive synchronized tracking control based on the cross-coupling technique which is first introduced to fulfill multi-robot assembly task in a centralized way by D. Sun et al. [35]. 
%   impose additional inter-coupling constraints between the networked robotic agents As to the multi-robot ensemble, consensus scheme of the networked robotic system with consideration of both uncertain dynamics and kinematics is investigated in [33]. To further enhance the system performance [36, 37], synchronization mechanism has been incorporated in the adaptive schemes to . Communication burden of this controller originated from the dependence on undirected cyclic information topology is later alleviated in .  

  A modified singularity robust technique proposed in~\cite{ren2017robust} can be adopted here to avoid potential singularity of the estimated kinematics and minimize the reconstruction error as far as possible.
      \begin{align}
 	\label{eq:41}
 	\begin{split}
    %  {\begin{array}{l}
\hat J_{e,i}^\dag ({\zeta _i},{{\hat \theta }_{k,i}}) &= \sum\nolimits_{k = 1}^{{n_{ns,i}}} {\frac{{{\sigma _{ik}}}}{{\sigma _{ik}^2 + \lambda _{Gik}^2}}{\nu _{ik}}\mu _{ik}^T} \\
{\lambda _{Gik}} &= {\lambda _{\max i}}\exp (-(\sigma _{ik}/{\Delta _i})^2)
% \end{array}}
\end{split}
    \end{align}
    where ${\sigma _{ik}}$ is the \emph{k}th singular value of the estimated Jacobian ${{\hat J_{e,i}}}$; ${\nu _{ik}}$ and ${\mu _{ik}}$ denote the \emph{k}th output and input singular vectors; ${n_{ns,i}}$ is the number of nonnull singular value of ${{\hat J_{e,i}}}$; ${\Delta _i}$ is the design constant which sets the size of the singularity region and ${\lambda _{\max i}}$ represents the maximum of the damping factor.
    
    Differentiating~\eqref{eq:40} with respect to time leads to the following reference acceleration
         \begin{align}
 	\label{eq:42}
     {\begin{array}{c}
{{\ddot \zeta }_{r,i}} = \hat J_{e,i}^\dag [{{\ddot{\hat x}}_{d,i}} - \sum\limits_{j \in {{{\cal N}}_i}} {{\varepsilon _i}(\Delta {{\dot x}_{o,i}} - \Delta {{\dot x}_{o,j}})}  - {\Lambda _i}{{\dot e}_i}] + \frac{{d{{\dot \zeta }_{sr,i}}}}{{dt}}\\
{\rm{ + }}\dot{\hat J}_{e,i}^\dag [{{\dot{\hat x}}_{d,i}} - \sum\limits_{j \in {{{\cal N}}_i}} {{\varepsilon _i}(\Delta {x_{o,i}} - \Delta {x_{o,j}})}  - {\Lambda _i}{e_i}]
\end{array}}
    \end{align} 
    where ${{\ddot{\hat x}_{d,i}}}$ represents the desired Cartesian-space acceleration of the \emph{i}th manipulator.
    
    Based on the reference velocity defined by (40), a joint-space sliding vectors ${s_i} \in {\Re ^{({n_r} - {n_c})}}$ is defined as follows
           \begin{align}
 	\label{eq:43}
     {{s_i} = {\dot \zeta _i} - {\dot \zeta _{r,i}}}
    \end{align}
    
    Incorporating~\eqref{eq:43} and~\eqref{eq:40} into~\eqref{eq:39} yields the following relation between the joint-space velocity and the Cartesian-space sliding vector
               \begin{align}
 	\label{eq:44}
     {{\hat J_{e,i}}{s_i} = {S_{x,i}} + {Y_{k,i}}{\tilde \theta _{k,i}} - {\dot{\tilde x}_{o,i}}}
    \end{align}
    
    Considering~\eqref{eq:21} and substituting~\eqref{eq:43} with its time derivative into~\eqref{eq:19} yields
    \begin{align}
 	\label{eq:45}
     {{M_{s,i}}{\dot s_i} + {C_{s,i}}{s_i} = {\tau _i} - J_{\phi ,i}^T{F_{I,i}} - {Y_{d,i}}({\zeta _i},{\dot \zeta _i},{\dot \zeta _{r,i}},{\ddot \zeta _{r,i}},{\beta _i}){\theta _{d,i}}}
    \end{align} 
    
    Now we propose the adaptive control law for the \emph{i}th mobile manipulator as
    \begin{align}
 	\label{eq:46}
 	\begin{split}
     {\tau _i} =&{\tau _{I,i}}+{Y_{d,i}}({\zeta _i},{\dot \zeta _i},{\dot \zeta _{r,i}},{\ddot \zeta _{r,i}},{\beta _i}){\hat \theta _{d,i}}\\
     &-({\hat J_{e,i}^T{K_{s,i}}{{\hat J}_{e,i}} + {K_{{\vartheta _i}}}}){s_i}
     \end{split}
    \end{align}
    where ${{K_{s,i}} \in \Re }$ and ${{K_{{\vartheta _i}}} \in \Re }$ are positive constants.
    
    The estimated dynamic and kinematic parameters ${\hat \theta _{d,i}}$, ${\hat \theta _{k,i}}$ are updated by
        \begin{align}
 	\label{eq:47}
     {{\dot{\hat \theta}_{d,i}} =  - {\Gamma _{d,i}}Y_{d,i}^T({\zeta _i},{\dot \zeta _i},{\dot \zeta _{r,i}},{\ddot \zeta _{r,i}},{\beta _i}){s_i}}
    \end{align}
    \begin{align}
     \label{eq:48}
     {{\dot{\hat \theta} _{k,i}} =  - {\alpha _i}{\Gamma _{k,i}}Y_{k,i}^T({\zeta _i},{\dot \zeta _i}){K_{o,i}}{\tilde x_{o,i}}}
    \end{align}
    where ${{K_{o,i}}}$ is a positive gain constant, ${{\Gamma _{d,i}}}$ and ${{\Gamma _{k,i}}}$ are positive definite matrices with opportune dimensions.
    
    The estimated parameters for the internal force regulation are updated by
       \begin{align}
     \label{eq:49}
     {{\dot{\hat \theta}_{Id,i}} =  - {\Gamma _{Id,i}}Y_{Id,i}^T({\zeta _i},{\dot \zeta _i},{F_{Id,i}}){s_i}}
    \end{align}
    where ${{\hat \theta _{Id,i}}}$ is the estimate of ${{\theta _{Id,i}}}$. ${{\Gamma _{Id,i}}}$ is a positive definite matrix with opportune dimension.
    \begin{remark}
    To avoid employing a separate step to excite the grasped object and estimate its dynamic and kinematic parameters with persistent exiting input signals before manipulating the unknown object~\cite{marino2018two}, the presented adaptive control in this section provides a more comprehensive approach to concurrently address the kinematic and dynamic uncertainties of both the grasped object and mobile manipulator and no persistent excitation condition is required.
    \end{remark}

	\section{Stability and Convergence Analysis}\label{sec:4}
	Two theorems will be given in detail in this section, which together illustrate the stability and error convergence of the proposed distributed adaptive cooperation scheme.
	
	We first define the parameter estimation error as
	   \begin{align}
     \label{eq:50}
     {{\tilde \theta _{sub}} = {\hat \theta _{sub}} - {\theta _{sub}}}
    \end{align}
    where the subscript $sub$ denotes the relevant linearized parameters defined above.
    
    To verify the efficacy of the distributed cooperative task allocation~\eqref{eq:28}, we define the first Lyapunov function candidate as
 	  \begin{align}
     \label{eq:51}
    %  {\begin{array}{c}
    \begin{split}
{V_d} = &\frac{1}{2}\varepsilon _d^T{\varepsilon _d} + \frac{1}{2}{\kappa _i}\sum\limits_{i = 1}^N {\left( {1 - {b_i}} \right){{{\cal P}}_i}\tilde \theta _{tr,i}^T\Gamma _{tr,i}^{ - 1}{{\tilde \theta }_{tr,i}}} \\
 = &\underbrace {\frac{1}{2}\varepsilon _{p,d}^T{\varepsilon _{p,d}} + \frac{1}{2}{\kappa _i}\sum\limits_{i = 1}^N {\left( {1 - {b_i}} \right){{{\cal P}}_i}\tilde \theta _{tr,p,i}^T\Gamma _{tr,i}^{ - 1}{{\tilde \theta }_{tr,p,i}}} }_{{V_{d,p}}}\\
 &+ \underbrace {\frac{1}{2}\varepsilon _{o,d}^T{\varepsilon _{o,d}} + \frac{1}{2}{\kappa _i}\sum\limits_{i = 1}^N {\left( {1 - {b_i}} \right){{{\cal P}}_i}\tilde \theta _{tr,o,i}^T\Gamma _{tr,i}^{ - 1}{{\tilde \theta }_{tr,o,i}}} }_{{V_{d,o}}}
% \end{array}}
\end{split}
    \end{align}   
where ${{\varepsilon _d} = {[\varepsilon _{d,1}^T,...,\varepsilon _{d,N}^T]^T} \in {\Re ^{Nm}}}$ is the collective error. ${{\varepsilon _{p,d}} = {[\varepsilon _{pd,1}^T,...,\varepsilon _{pd,N}^T]^T}}$ and ${{\varepsilon _{o,d}} = {[\varepsilon _{od,1}^T,...,\varepsilon _{od,N}^T]^T}}$ are both $3N \times 1$ vectors. ${{\tilde \theta _{tr,i}} = {\hat \theta _{tr,i}} - {\theta _{tr,i}}}$ is the estimation error of ${{\theta _{tr,i}}}$.

The first time derivative of ${{V_d}}$ can be expressed as:
       \begin{align}
     \label{eq:52}
     {{\dot V_d} = \varepsilon _d^T{\dot \varepsilon _d} + {\kappa _i}\sum\limits_{i = 1}^N {\left( {1 - {b_i}} \right){{{\cal P}}_i}\tilde \theta _{tr,i}^T\Gamma _{tr,i}^{ - 1}{{\dot{\hat \theta}}_{tr,i}}}}
    \end{align}
    
    Combining~\eqref{eq:24} with~\eqref{eq:25} yields
        \begin{align}
     \label{eq:53}
     {{\varepsilon _{d,i}} = {\delta _i} - \left( {1 - {b_i}} \right)\left[ {\left( {{I_6} \otimes {{{\cal F}}_{tr}}\left( t \right)} \right){{\tilde \theta }_{tr,i}} + {{{\tilde {\cal T}}}_{ti}}} \right]}
    \end{align}   
    where ${{{\tilde {\cal T}}_{ti}} = {{\hat {\cal T}}_{ti}} - {{{\cal T}}_{ti}}}$.
    
    Differentiating~\eqref{eq:24} and considering~\eqref{eq:28} leads to
            \begin{align}
     \label{eq:54}
     {{\dot \varepsilon _{d,i}} =  - \kappa {{{\cal P}}_i}{\gamma _i}}
    \end{align} 
    
    Incorporating the above two equations,~\eqref{eq:27} and~\eqref{eq:29} into~\eqref{eq:52} gives
        \begin{align}
        \begin{split}
     \label{eq:55}
%     {\begin{array}{c}
{{\dot V}_d} =&- \kappa {\delta ^T} \left[ {{{\cal P}}\left( {{{\cal L}} + {{\cal B}}} \right) \otimes {I_6}} \right] \delta  + \kappa \sum\limits_{i = 1}^N {{{{\cal P}}_i}(1 - {b_i}){{\bar \rho }_i}} \\
 =&- \frac{\kappa }{2}{\delta ^T}\left[ {\left( {{{\cal P}}\left( {{{\cal L}} + {{\cal B}}} \right) + {{\left( {{{\cal L}} + {{\cal B}}} \right)}^T}{{\cal P}}} \right) \otimes {I_6}} \right]\delta\\  
 &+ \kappa \sum\limits_{i = 1}^N {{{{\cal P}}_i}(1 - {b_i}){{\bar \rho }_i}} \\
 =&\underbrace { - \frac{\kappa }{2}\delta _p^T\left( {{{\cal Q}} \otimes {I_3}} \right){\delta _p} + \kappa \sum\limits_{i = 1}^N {{{{\cal P}}_i}(1 - {b_i}){{\bar \rho }_i}} }_{{{\dot V}_{d,p}}} \\
 &\underbrace { - \frac{\kappa }{2}\delta _o^T\left( {{{\cal Q}} \otimes {I_3}} \right){\delta _o}}_{{{\dot V}_{d,o}} \le 0}
%\end{array}}
\end{split}
    \end{align} 
    where ${\bar \rho _i} = \delta _i^T\sum\nolimits_{j \in {{{\cal N}}_i}} {{{{\tilde {\cal T}}}_{ji}}}  + \gamma _i^T{{\tilde {\cal T}}_{ti}}$ and Lemma 1 are employed. ${\delta  = {[\delta _1^T,...,\delta _N^T]^T}}$ denotes the concatenation of estimation error ${\delta _i}$ of all agents. ${{{\cal Q}}}$ is positive definite matrix defined in Lemma 2. ${{\delta _o} = {[\delta _{o,1}^T,...,\delta _{o,N}^T]^T} \in {\Re ^{3N}}}$ and ${{\delta _p} = {[\delta _{p,1}^T,...,\delta _{p,N}^T]^T} \in {\Re ^{3N}}}$ are two rearranged components of ${\delta }$.
    
    With~\eqref{eq:51} and~\eqref{eq:52}, we are ready to give the following theory:
    
    \emph{Theorem 1}: As for the cooperative task allocation to the   mobile manipulators, the proposed distributed estimation law~\eqref{eq:28} with the trajectory parameter updating law~\eqref{eq:29} guarantee the estimation error convergence of desired allocated task trajectory, i.e. ${{\delta _i} = {\hat x_{d,i}} - {{{\cal T}}_{ti}} - {x_{td}} \to 0}$ and ${{\dot \delta _i} \to 0}$. 
    
    	\begin{proof}
	    Please see Appendix B for the proof.	
	\end{proof}
	
	To verify the convergence of the distributed synchronization controller~\eqref{eq:46}, the control law~\eqref{eq:46} with~\eqref{eq:34} are incorporated into~\eqref{eq:45} to establish the following closed-loop dynamic equation as:
           \begin{align}
           \begin{split}
     \label{eq:56}
%     {\begin{array}{c}
{M_{s,i}}{{\dot s}_i}&+{C_{s,i}}{s_i} ={Y_{d,i}}({\zeta _i},{{\dot \zeta }_i},{{\dot \zeta }_{r,i}},{{\ddot \zeta }_{r,i}},{\beta _i}){{\tilde \theta }_{d,i}} - J_{\phi ,i}^T{F_{I,i}} \\
 &+ \hat J_{\phi ,i}^T{F_{Id,i}}- \left[ {\hat J_{e,i}^T({k_{r,i}} + {K_{s,i}}){{\hat J}_{e,i}} + {K_{{\vartheta _i}}}} \right]{s_i}
%\end{array}}
\end{split}
    \end{align}
    
    Then, the following Lyapunov function candidate ${V}= \sum\nolimits_{i = 1}^N {{V_i}} $ is designed
 \begin{align}   
\label{eq:57}
\begin{split}
V =&\sum\nolimits_{i = 1}^N {\left[ \frac{1}{2}{s_i^T{M_{s,i}}{s_i} + e_i^T\left( {1 - {1 \mathord{\left/
 {\vphantom {1 {{K_\varepsilon }}}} \right.
 \kern-\nulldelimiterspace} {{K_\varepsilon }}}} \right){K_{s,i}}{\Lambda _i}{e_i}} \right]} \\
&+\sum\nolimits_{i = 1}^N {\left(\frac{1}{2}{\tilde \theta _{Id,i}^T\Gamma _{Id,i}^{ - 1}{{\tilde \theta }_{Id,i}} +\frac{1}{2}\tilde x_{o,i}^T{\alpha _i}{K_{o,i}}{{\tilde x}_{o,i}}} \right)} \\
&+\sum\nolimits_{i = 1}^N {\left( \frac{1}{2}{\tilde \theta _{d,i}^T\Gamma _{d,i}^{ - 1}{{\tilde \theta }_{d,i}} + \frac{1}{2}\tilde \theta _{k,i}^T\Gamma _{k,i}^{ - 1}{{\tilde \theta }_{k,i}}} \right)}
 \end{split}
\end{align}
 where ${{K_\varepsilon }}$ is a positive constant chosen as ${{K_\varepsilon } > 1}$.
 
 Differentiating ${V_i}$ with respect to time leads to
  \begin{align}   
         \label{eq:58}
         \begin{split}
    %  {\begin{array}{c}
    {{\dot V}_i}= &s_i^T\left[{M_{s,i}}{{\dot s}_i}+ {({{\dot M}_{s,i}}{s_i})}/2\right]+ \tilde x_{o,i}^T{\alpha _i}{K_{o,i}}{{\dot{\tilde x}}_{o,i}} \\
&+ \tilde \theta _{d,i}^T\Gamma _{d,i}^{ - 1}{{\dot{\tilde \theta}}_{d,i}}+ \tilde \theta _{Id,i}^T\Gamma _{Id,i}^{ - 1}{{\dot{\tilde \theta}}_{Id,i}} + \tilde \theta _{k,i}^T\Gamma _{k,i}^{ - 1}{{\dot{\tilde \theta}}_{k,i}}\\
    &+2e_i^T\left( {1 - 1/K_\varepsilon }\right){K_{s,i}}{\Lambda _i}{{\dot e}_i}
    \end{split}
    \end{align}
    
    Incorporating~\eqref{eq:56} into~\eqref{eq:58} with Property 2 yields
\begin{align}   
         \label{eq:59}
         \begin{split}
{{\dot V}_i} =& s_i^T\left[ {{Y_{d,i}}({\zeta _i},{{\dot \zeta }_i},{{\dot \zeta }_{r,i}},{{\ddot \zeta }_{r,i}},{\beta _i}){{\tilde \theta }_{d,i}} + {{({{\dot \beta }_i}J_{\phi ,i}^T{M_o}{J_{\phi ,i}}{s_i})}/2}} \right]\\
 &- s_i^T\left[ {\hat J_{e,i}^T({K_{s,i}} + {k_{r,i}}){{\hat J}_{e,i}} + {K_{{\vartheta _i}}}} \right]{s_i} + \tilde x_{o,i}^T{\alpha _i}{K_{o,i}}{{\dot{\tilde x}}_{o,i}}\\
 &+ s_i^T( {\hat J_{\phi ,i}^T{F_{Id,i}} - J_{\phi ,i}^T{F_{I,i}}}) + 2e_i^T\left( {1 - {1/{K_\varepsilon }}} \right){K_{s,i}}{\Lambda _i}{{\dot e}_i}\\
&+ \tilde \theta _{Id,i}^T\Gamma _{Id,i}^{ - 1}{{\dot{\tilde \theta}}_{Id,i}}+ \tilde \theta _{k,i}^T\Gamma _{k,i}^{ - 1}{{\dot{\tilde \theta}}_{k,i}} + \tilde \theta _{d,i}^T\Gamma _{d,i}^{ - 1}{{\dot{\tilde \theta}}_{d,i}}
\end{split}
\end{align}
    in which the term ${s_i^T( {\hat J_{\phi ,i}^T{F_{Id,i}} - J_{\phi ,i}^T{F_{I,i}}})}$ can be reformulated as:
  \begin{align}   
         \label{eq:60}
{s_i^T\left( {\hat J_{\phi ,i}^T{F_{Id,i}} - J_{\phi ,i}^T{F_{I,i}}} \right) = s_i^TJ_{\phi ,i}^T{\tilde F_{I,i}} + s_i^T{Y_{Id,i}}{\tilde \theta _{Id,i}}}
    \end{align} 
    where ${{\tilde F_{I,i}} = {F_{Id,i}} - {F_{I,i}}}$ is the internal force error.
    
    Folding~\eqref{eq:37}, ~\eqref{eq:44} and the updating laws~\eqref{eq:47}-\eqref{eq:49} into ~\eqref{eq:59} gives
    \begin{align}   
         \label{eq:61}
\begin{split}
{{\dot V}_i} \le& - \left( {1 - \frac{1}{{{K_\varepsilon }}}} \right)S_{x,i}^T{K_{s,i}}{S_{x,i}} + 2e_i^T\left( {1 - \frac{1}{{{K_\varepsilon }}}} \right){K_{s,i}}{\Lambda _i}{{\dot e}_i}\\
&- \alpha _i^2\left( {{K_{s,i}} + {K_{o,i}} - {K_\varepsilon }{K_{s,i}}} \right)\tilde x_{o,i}^T{{\tilde x}_{o,i}} - {k_{r,i}}{\left\| {{{\hat J}_{e,i}}{s_i}} \right\|^2}\\
&+ s_i^TJ_{\phi ,i}^T{{\tilde F}_{I,i}}- s_i^T\left( {{K_{{\vartheta _i}}} - {\vartheta _i}} \right){s_i}
\end{split}
    \end{align} 
       whose derivation detail is attached in Appendix C. and the following the inequality is used
    \begin{align}   
         \label{eq:62}
{- 2{\alpha _i}{K_{s,i}}S_{x,i}^T{\tilde x_{o,i}} \le \frac{1}{{{K_\varepsilon }}}S_{x,i}^T{K_{s,i}}{S_{x,i}} + \alpha _i^2{K_\varepsilon }{K_{s,i}}\tilde x_{o,i}^T{\tilde x_{o,i}}}
    \end{align}

    The last but one term on the right side of~\eqref{eq:61} satisfies:
      \begin{align}   
         \label{eq:63}
\begin{split}
&\sum\nolimits_{i = 1}^N {s_i^TJ_{\phi ,i}^T{{\tilde F}_{I,i}}}  = \sum\nolimits_{i = 1}^N {{{\left( {{J_{e,i}}{s_i}} \right)}^T}{{\left( {J_{o,i}^\dag } \right)}^T}{{\tilde F}_{I,i}}} \\
 \le & \sum\nolimits_{i = 1}^N {\left\| {{J_{e,i}}{s_i}} \right\|\left( {\left\| {{{\bar F}_{Id,i}}} \right\| + \left\| {{{\bar F}_{s,i}}} \right\|} \right)} 
\end{split}
    \end{align} 
    where the following inequality is employed
      \begin{align}   
         \label{eq:64}
\begin{split}
{\left( {J_{o,i}^\dag } \right)^T}{{\tilde F}_{I,i}} = &{\left( {J_{o,i}^\dag } \right)^T}\left( {{F_{Id,i}} - {F_{I,i}}} \right)\\
 =&{{\bar F}_{Id,i}} - {{\bar F}_{I,i}}\\
 \le&\left\| {{{\bar F}_{Id,i}}} \right\| + \left\| {{{\bar F}_{s,i}}} \right\|
\end{split}
    \end{align}
    
    Considering~\eqref{eq:39} and~\eqref{eq:63}, ~\eqref{eq:61} can be further simplified into
 \begin{align}   
         \label{eq:65}
\begin{split}
{{\dot V}_i} \le&- \alpha _i^2\left( {{K_{s,i}} + {K_{o,i}} - {K_\varepsilon }{K_{s,i}}} \right)\tilde x_{o,i}^T{{\tilde x}_{o,i}} - s_i^T\left( {{K_{{\vartheta _i}}} - {\vartheta _i}} \right){s_i}\\
&- \left( {1 - \frac{1}{{{K_\varepsilon }}}} \right){K_{s,i}}\left( {\dot e_i^T{{\dot e}_i} + e_i^T\Lambda _i^T{\Lambda _i}{e_i}} \right)\\
 \le& 0
\end{split}
    \end{align}   
    where the positive constants ${{K_{{\vartheta _i}}}}$, ${{K_\varepsilon }}$, ${{K_{o,i}}}$, ${{k_{r,i}}}$ are set as
\begin{align}   
         \label{eq:66}
{\left\{ \begin{array}{l}
{K_{{\vartheta _i}}} > {\vartheta _i}\\
{K_\varepsilon } > 1\\
{K_{oi}} > \left( {{K_\varepsilon } - 1} \right){K_{s,i}}\\
{k_{r,i}} \ge {{\left\| {{J_{e,i}}{s_i}} \right\|\left( {\left\| {{{\bar F}_{Id,i}}} \right\| + \left\| {{{\bar F}_{s,i}}} \right\|} \right)}/{{{| {{{\hat J}_{e,i}}{s_i}} \|}^2}}}
\end{array} \right.}
    \end{align} 
    
    Now we are in position to formulate the following theorem.
    
    \emph{Theorem 2}: For the $N$ uncertain mobile manipulators cooperatively grasping an unknown object, the distributed adaptive control law~\eqref{eq:46} with parameter updating laws~\eqref{eq:47}-\eqref{eq:49} can guarantee the stability of the robotic ensemble and lead to the motion synchronization and the convergence of Cartesian-space motion tracking errors under the condition of directed spanning tree, i.e. ${\Delta {x_{e,i}},{\rm{ }}\Delta {\dot x_{e,i}} \to 0}$, ${\Delta {x_{e,i}} - \Delta {x_{e,j}} \to 0}$ and ${\Delta {\dot x_{e,i}} - \Delta {\dot x_{e,j}} \to 0}$ as $t \to \infty $, $\forall i \in \{ 1,2,...,N\} $. Furthermore, the internal force tracking error exponentially approaches to the neighborhood of the zero point.
    \begin{proof}
    Please see Appendix C for the proof.
    \end{proof}

	\section{Validation Results}\label{sec:vr}
\begin{figure}[ht]
		\sf
		\centering
		\normalsize
		\includegraphics[width=0.45\textwidth]{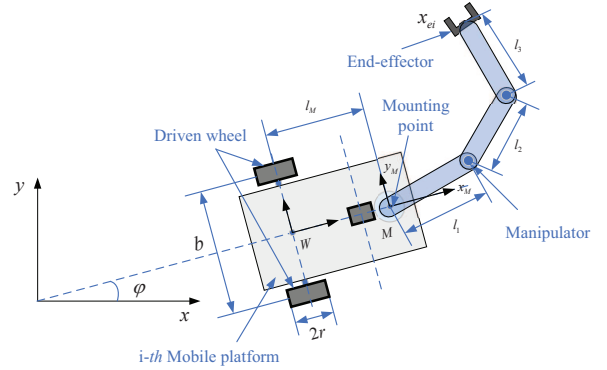}
		\caption{Schematic diagram of the mobile manipulator employed in the simulation\label{figure3}}
	\end{figure}

\begin{figure}[h]
		\sf
		\centering
		\normalsize
		\includegraphics[width=0.5\textwidth]{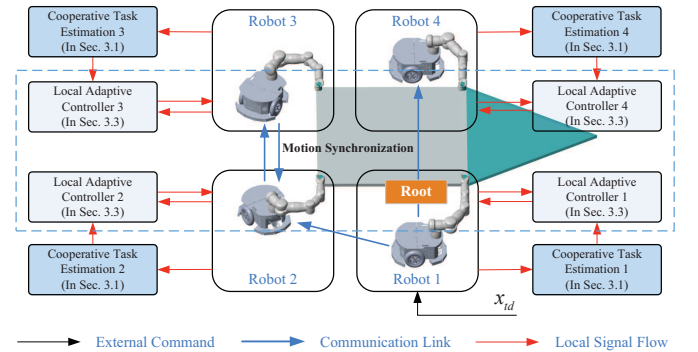}
		\caption{Communication topology of the networked mobile manipulators\label{figure4}}
	\end{figure}

In this section, we present the simulation results to validate the efficacy and demonstrate the performance of the proposed distributed adaptive control scheme using four nonholonomic mobile manipulators whose motions are constrained in the horizontal X-Y plane. The schematic diagram of the NMM and the communication graph between the four networked NMMs are shown in Fig.~\ref{figure3} and Fig.~\ref{figure4}. Dynamic and kinematic parameters of the mobile manipulator are listed in Table A-\uppercase\expandafter{\romannumeral1} in Appendix D. Three cases are studied to give the readers a deep insight into the interaction between the complex interconnected mechanism, the distributed manner and the synchronization considered in this work. These simulations are implemented by employing Simulink and SimMechanics 2G. Nonholonomic constraints imposed on the mobile platform are simulated based on the Lagrange Equation and Lagrange multiplier method. In both cases, the mobile manipulators only exchange the estimated cooperative task position ${{\hat x_{d,i}}}$ and the observed Cartesian-space position ${{x_{o,i}}}$ with their neighbors according to the communication topology of a directed spanning tree, as implied by the mathematical expression of the proposed scheme.
The desired trajectory of the virtual leader is supposed to be known only to Robot 1 (Please see Figure 4 for the robot number). Then the Laplacian matrix ${{\cal L}}$ and the matrix ${{\cal B}}$ with the vector ${w}$ defined in Lemma 1 and Lemma 2 associated with the communication graph is: ${w = {[1,3,4,2]^T}}$\\
\begin{displaymath}
{{\cal L}} = \left[ {\begin{array}{*{20}{c}}
0&0&0&0\\
{ - 1}&2&{ - 1}&0\\
0&{ - 1}&1&0\\
{ - 1}&0&0&1
\end{array}} \right],\quad 
{{\cal B}} = \left[ {\begin{array}{*{20}{c}}
1&0&0&0\\
0&0&0&0\\
0&0&0&0\\
0&0&0&0
\end{array}} \right].
\end{displaymath}

The initial position derivation of the EE of Robot1 ${{r_{01}} = {[1, - 1,0]^T}}$ and the relative position between the EEs of the mobile manipulators are set as ${{r_{12}} = {[ - 2,0,0]^T}}$, ${r_{13}} = {[ - 2,2,0]^T}$, ${{r_{14}} = {[0,2,0]^T}}$.
	
\begin{figure*}[ht]
\centering
  \includegraphics[width=\textwidth]{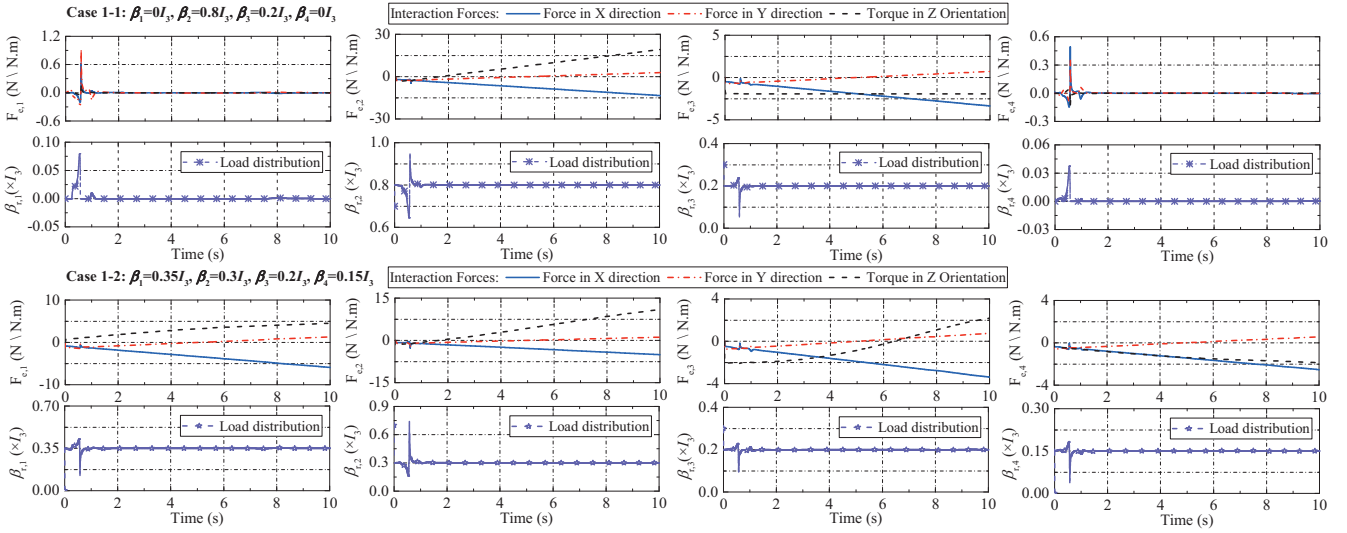}
\caption{Interaction forces and load sharing under two different load distribution scheme}
\label{case1}
\end{figure*}
	
\subsection{Interaction of the interconnected system}
This case study presents the results under two different desired load distribution schemes, which aims at validating the feasibility of the distributed dynamic model of the whole interconnected system, as given in~\eqref{eq:19}. To highlight the relation between resultant interaction forces and designated load distribution in~\eqref{eq:16}, the system dynamics and kinematics here are assumed to be certain and the cooperative task is exactly allocated to each NMM. Desired cooperative trajectory is selected as ${x_{td}} = {[{t^3}/25+{t^2}/5,-t^3/50+3t^2/10,-\pi t^3/3000+\pi t^2/200]^T}$.

Fig.~\ref{case1} presents the interaction forces ${F_{e,i}}$ and actual load sharing ${\beta_{r,i}}$ during the cooperative transport under two different load distribution schemes, i.e. case 1-1: ${\beta_1 = \beta_4=0I_3}$, ${\beta_2=0.8I_3}$, ${\beta_3 = 0.2I_3}$ and case 1-2: ${\beta_1=0.35I_3}$, ${\beta_2=0.3I_3}$, ${\beta_3=0.2I_3}$, ${\beta_4=0.15I_3}$. Here ${\beta_{r,i}}$ is computed based on the physically plausible wrench decomposition method proposed in~\cite{donner2018physically}. After the interconnected system achieve controlled dynamic balance, approximately after $t=1$ s, the actual load sharing parameters ${\beta_{r,i}}$ match the initially designated desired load distribution parameters ${\beta_i}$ well. For those NMMs which are set to not share any load (Robot 1 and Robot 4 in case 1), the interaction forces between their EEs and the tightly grasped object approxiamte to zero.
\begin{figure*}[ht]
\centering
  \centering
  \includegraphics[width=0.98\textwidth]{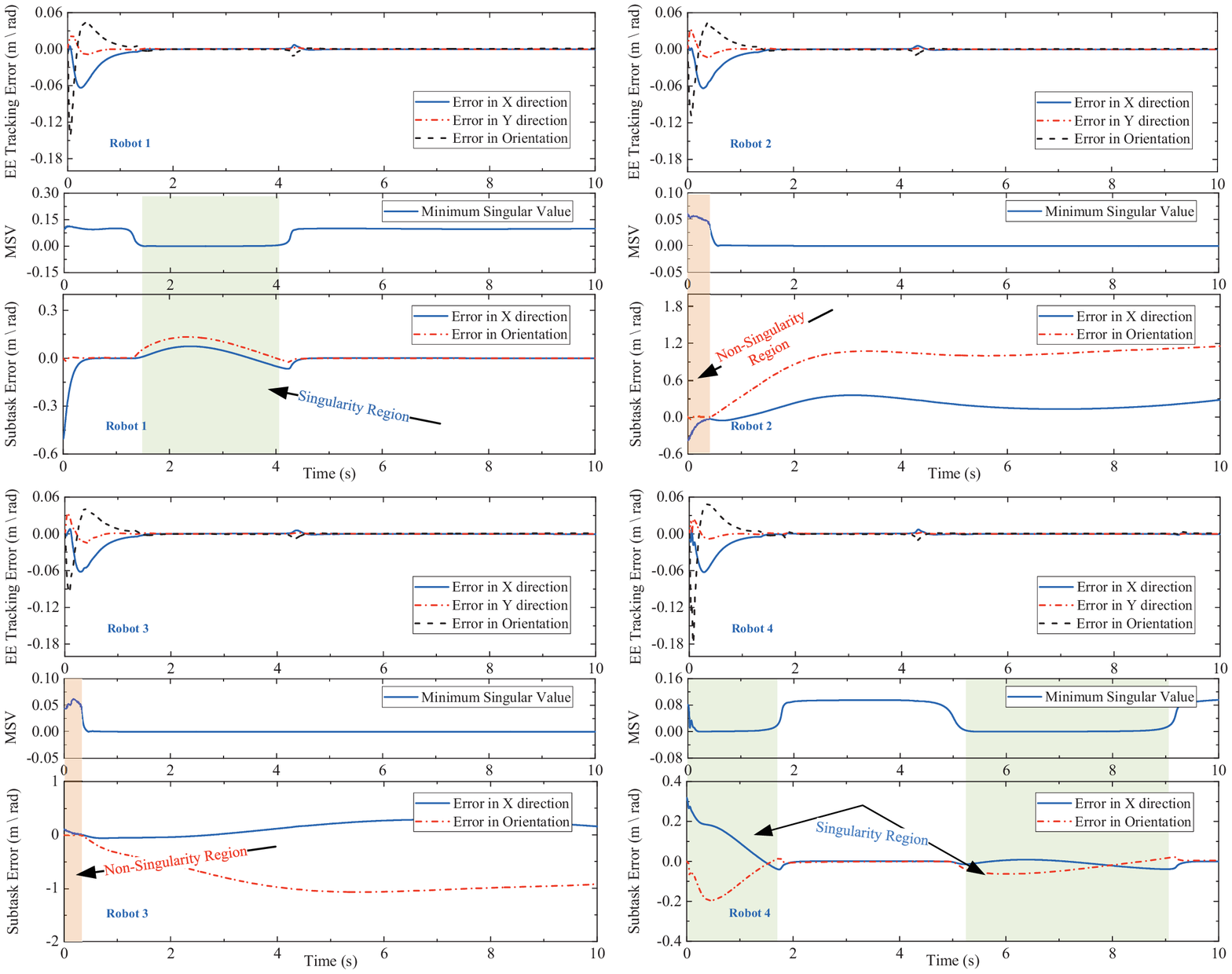}
\caption{Primary and sub task tracking errors of the four NMMs}
\label{figure5}
\end{figure*}
\subsection{Distributed kinematic cooperation}
In this case, four networked NMMs are controlled in a distributed way and their end-effector motions are coordinated so that the equivalent centroid of the kinematic cooperation task (ECCT, denoted by ${{x_{ecct}} = {[{x_{ec}},{y_{ec}},{\alpha _{ec}}]^T} \in {\Re ^3}}$ and visualized by blue triangle in the supplementary video) can track the desired trajectory of the virtual leader (DTVL, denoted by ${x_{dtvl}} \in {\Re ^3}$ and visualized by red triangle in the supplementary video). Inspired by the definition of the centroid variable in~\cite{basile2013decentralized} and to maintain the consistency between Case 2 and Case 3, the definition of ${{x_{ecct}}}$ is given as ${{x_{ecct}} = \sum\nolimits_{i = 1}^4 {{x_{e,i}}/4}
  + R({\alpha _{ec}}){r_{ec}}}$, where ${R({\alpha _{ec}}) = [\cos ({\alpha _{ec}}), - \sin ({\alpha _{ec}});\sin ({\alpha _{ec}}),\cos ({\alpha _{ec}})]}$ denotes the rotation matrix associated with the rotation angle of ECCT and ${{r_{ec}} = {[1.04 \times (\sqrt 3  + 1),0]^T}}$ denotes the virtual link between ECCT and the kinematic centroid of the four frames attached to the end-effectors. The DTVL is chosen as\\
 \begin{displaymath}
 {{x_{dtvl}} = \left[ \begin{array}{l}
1.04(\sqrt 3  + 1) + 0.1 + {{3{t^2}} \mathord{\left/
 {\vphantom {{3{t^2}} {20}}} \right.
 \kern-\nulldelimiterspace} {20}} - {{{t^3}} \mathord{\left/
 {\vphantom {{{t^3}} {100}}} \right.
 \kern-\nulldelimiterspace} {100}}\\
0.1 + 0.3\sin (0.1\pi t) + 0.1\cos (0.1\pi t)\\
 \qquad+ 0.3\sin (0.2\pi t) - 0.1\cos (0.2\pi t)\\
{\pi  \mathord{\left/
 {\vphantom {\pi  {15}}} \right.
 \kern-\nulldelimiterspace} {15}} - {{\pi {t^2}} \mathord{\left/
 {\vphantom {{\pi {t^2}} {500}}} \right.
 \kern-\nulldelimiterspace} {500}} + {{\pi {t^3}} \mathord{\left/
 {\vphantom {{\pi {t^3}} {7500}}} \right.
 \kern-\nulldelimiterspace} {7500}}
\end{array} \right]}
\end{displaymath}	
which can be linearized according to~\eqref{eq:23}. The allocated task of this cooperative task to each robot agent are taken as the primary task.
The local subtask for each NMM ${x_{sd,i}} = {[{x_{M,i}},{\varphi _{M,i}}]^T}$ is set as
\begin{displaymath}
 {x_{sd,i}} = \left[ \begin{array}{l}
0.1 + {{3{t^2}} \mathord{\left/
 {\vphantom {{3{t^2}} {20}}} \right.
 \kern-\nulldelimiterspace} {20}} - {{{t^3}} \mathord{\left/
 {\vphantom {{{t^3}} {100}}} \right.
 \kern-\nulldelimiterspace} {100}}\\
{{\pi {t^2}} \mathord{\left/
 {\vphantom {{\pi {t^2}} {25}}} \right.
 \kern-\nulldelimiterspace} {25}} - 2{{\pi {t^3}} \mathord{\left/
 {\vphantom {{\pi {t^3}} {375}}} \right.
 \kern-\nulldelimiterspace} {375}}\underbrace { \mp {\pi  \mathord{\left/
 {\vphantom {\pi  3}} \right.
 \kern-\nulldelimiterspace} 3}}_{for{\rm{  }}i = 2,3}
\end{array} \right]
\end{displaymath}
 where ${x_{M,i}}$ and ${\varphi _{M,i}}$ denote the x-coordinate and the orientation of the mobile base.

   The internal force control~\eqref{eq:34} and the object dynamics part in the synthesized dynamic equation~\eqref{eq:19} are set to zero. The uncertainties of the dynamic/kinematic/trajectory parameters known to each NMM are listed in Table A-\uppercase\expandafter{\romannumeral2} in Appendix~D. Here we use the 2-norm to quantify the discrepancy between the real value and the initial values (uncertainty) of these parameters. The dynamic regressor ${{Y_{d,i}}({\zeta _i},{\dot \zeta _i},{\ddot \zeta _i},0)}$ is a ${5 \times 74}$ matrix and the kinematic regressor ${{Y_{k,i}}({\zeta _i},{\dot \zeta _i})}$ is a $3 \times 9$ matrix, whose analytical expressions are not presented here. The parameters for the whole adaptive scheme are selected as presented in Table A-\uppercase\expandafter{\romannumeral3} in Appendix~D.

Fig.~\ref{figure5} shows the tracking results of the primary task and subtask for each robot, from which one can conclude that the local subtask tracking performance is partly sacrificed to maintain the high tracking performance of the primary task when the projected Jacobian is singular. In non-singularity region, both two tasks are fully executed. It should be noted that transition to non-/singularity region during the task execution would introduce impact due to the sudden change in the reference joint trajectory. This phenomenon can be alleviated by adjusting the singularity region parameter ${\Delta _i}$ and the maximum of the damping factor ${\lambda _{\max i}}$ defined in~\eqref{eq:41}.

\begin{figure*}[ht]
		\sf
		\centering
		\normalsize
		\includegraphics[width=\textwidth]{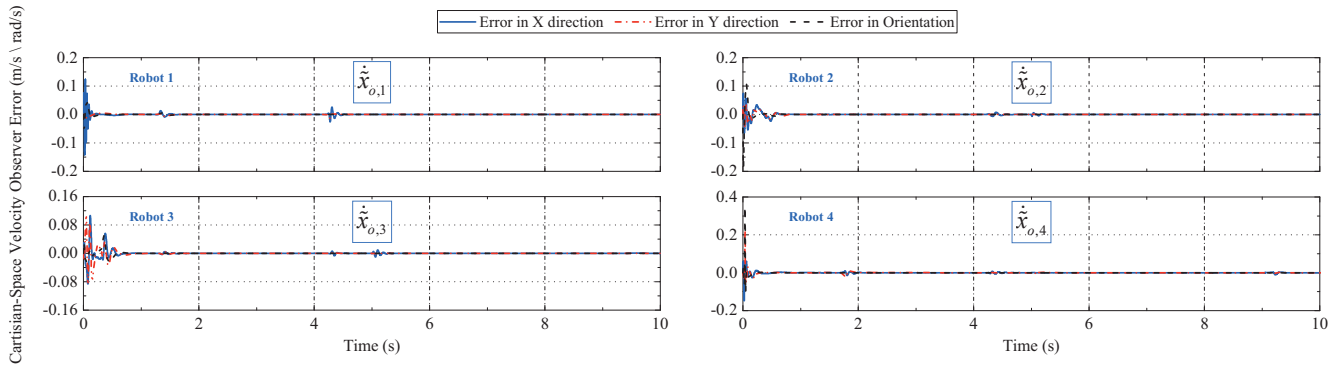}
		\caption{Cartesian-space velocity observer error of the four NMMs}
		\label{figure6}
	\end{figure*}

\begin{figure*}
\centering
\begin{subfigure}{.48\textwidth}
  \centering
  \includegraphics[width=\textwidth]{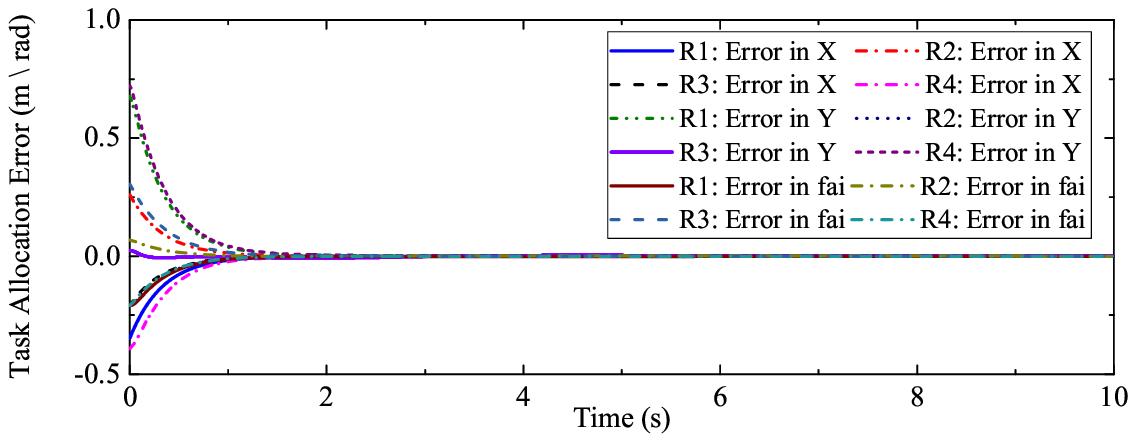}
  \caption{Task allocation error}
  \label{fig:sub7a}
\end{subfigure}%
\begin{subfigure}{.48\textwidth}
  \centering
  \includegraphics[width=\textwidth]{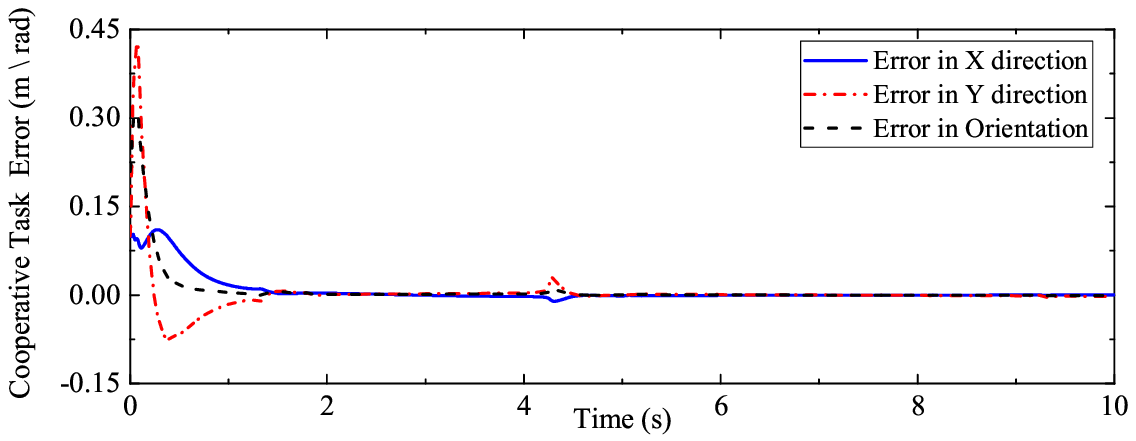}
  \caption{Cooperative task tracking error}
  \label{fig:sub7b}
\end{subfigure}
\caption{Cooperative task tracking error and task allocation error}
\label{figure7}
\end{figure*}

\begin{figure}
\centering
  \centering
  \includegraphics[width=0.48\textwidth]{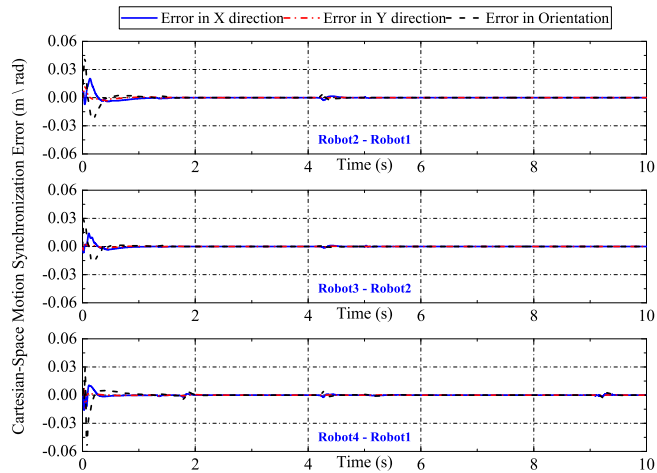}
\caption{Synchronization errors between the four NMMs}
\label{figure8}
\end{figure}

In Fig.~\ref{figure6}, Cartesian-space velocity observer errors by the distributed observer~\eqref{eq:35} are shown to converge to zero. The convergence of synchronization errors between the Cartesian-space motions of the mobile manipulator illustrated in Fig.~\ref{figure8} and the convergence of distributed task allocation error in Fig.~\ref{figure7} a) together contribute to the convergence of the cooperative task tracking error presented in Fig.~\ref{figure7} b), as expected by Theorem 2.

\subsection{Distributed cooperative transport}
In this case, the desired internal force is set to zero due to the tight connection condition.
To show the superiority of the proposed distributed adaptive control scheme (DA scheme) in the cooperative task tracking and internal force regulation, a conventional visual servoing control scheme (NA scheme) without adaptation is introduced. These two comparative schemes share the same control gains, initial values and communication topology and both have the access to the pose feedback signals. The desired trajectory for the operational point of the grasped object is given as ${x_{td}={[{t^3}/25+{t^2}/5,-t^3/50+3t^2/10,0]^T}}$. Dynamic parameters of the grasped object and kinematic parameters associated with the grasp matrix are listed in Table A-\uppercase\expandafter{\romannumeral2} and unknown to the robotic agent. The task allocation parameters and most of the control gains are selected as in case 2, while ${{\varepsilon _i} = 5}$, ${\Delta _i}{\rm{ = }}0.08$, ${\lambda _{\max i}}{\rm{ = 0}}{\rm{.15}}$ and ${\Gamma _{d,i}}=5{I_3}$. The desired load distribution to each NMM is set as: ${\beta _1} = 0.5{I_3}$, ${{\beta _2} = 0.5{I_3}}$, ${\beta _3} = {\beta _4} = {0_{3 \times 3}}$.
 
\begin{figure*}
\centering
\begin{subfigure}{.5\textwidth}
  \centering
  \includegraphics[width=1\linewidth]{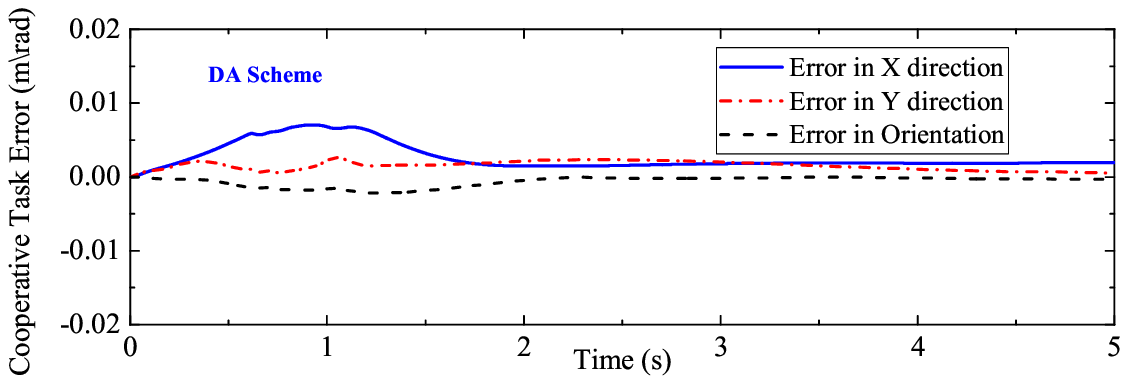}
  \caption{Tracking error with proposed DA scheme}
  \label{fig:sub9a}
\end{subfigure}%
\begin{subfigure}{.5\textwidth}
  \centering
  \includegraphics[width=1\linewidth]{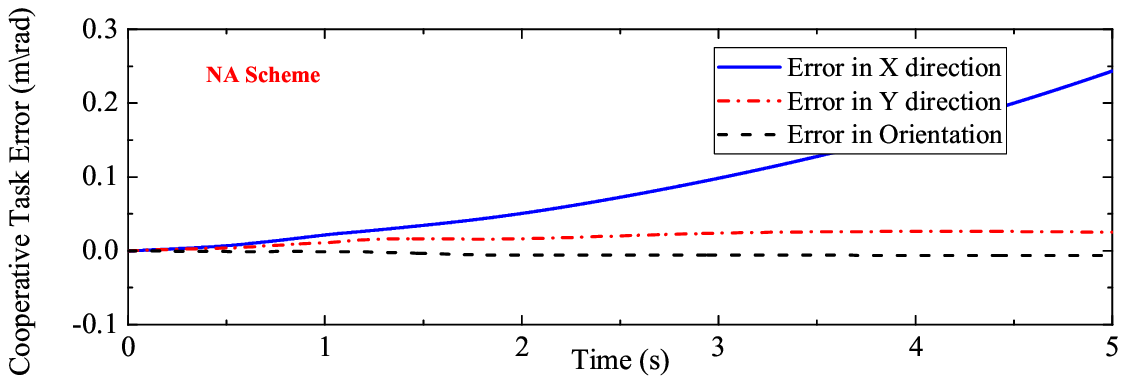}
  \caption{Tracking error with NA scheme}
  \label{fig:sub9b}
\end{subfigure}
\caption{Cooperative transport error with two comparative schemes}
\label{figure9}
\end{figure*}

 The tracking errors of operational point in this cooperative transport task are presented in Fig.~\ref{figure9}. Maximum and root mean square (RMS) of the 2-norm of the tracking error are listed in Table \uppercase\expandafter{\romannumeral1}, which implies that the proposed distributed adaptive cooperation scheme significantly improves the tracking performance in the cooperative transport task even in the presence of uncertain dynamics/kinematics and constrained communication.

\begin{figure*}[ht]
		\sf
		\centering
		\normalsize
		\includegraphics[width=0.96\textwidth]{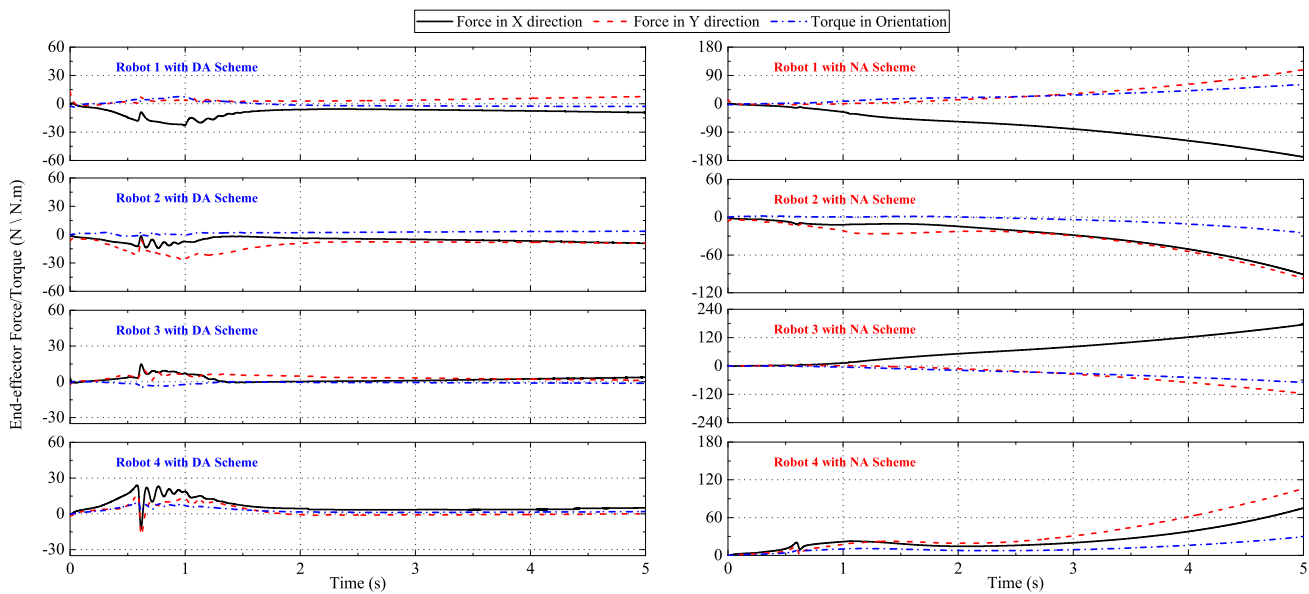}
		\caption{Interaction forces of the four mobile manipulators with two comparative schemes
		\label{figure10}}
	\end{figure*}

In addition, Fig.~\ref{figure10} displays the measured interaction forces between the EEs of the mobile manipulators and the grasped object during the task execution with the two schemes, from which one can easily concludes that the proposed adaptive scheme enjoys much smaller interaction forces than conventional NA scheme. According to the constraints defined in~\eqref{eq:30}, smaller interaction forces imply smaller internal forces and further better transient performance. This advantage owes to the synchronization of the allocated tasks achieved by the distributed cooperative task allocation in Section 3.1 and the motion synchronization achieved by the distributed adaptive control in Section 3.3.

Different from the results in other related papers that the internal force is presented to be very large in the presence of small kinematic discrepancy, here we employ very small control gains in the simulations to highlight the improvement of the tracking performance with the proposed adaptive scheme.

In practical implementation, requirement of EE force/torque sensors can be avoided at the expense of small performance degradation. Then the adaptively regulated gain ${{k_{r,i}}}$ can be given as ${{k_{r,i}} \ge {k_{rc}}{\|J_{e,i}{s_i}\|}/{\|\hat J}_{e,i}{s_i}\|^2}$, where ${{k_{rc}}}$ is a positive constant. Another advantage this reformulation has is that the stability of the system can be more easily guaranteed because there exists an unavoidable lag in the force/torques signals feedback to the calculation of ${{k_{r,i}}}$ due to the mechanical causality in real application.
\begin{table}[h]
\caption{THE PERFORMANCE INDEXES WITH TWO COMPARATIVE SCHEMES} 
\begin{tabular*}{0.48\textwidth}{c@{\extracolsep{\fill}}*{4}{c}}
\cmidrule{1-5}\morecmidrules\cmidrule{1-5}
    &\multicolumn{2}{c}{Proposed adaptive scheme} &\multicolumn{2}{c}{Non-adaptive scheme}\\
                     &Max  &RMS                   &Max  &RMS \\
\midrule
 $\Delta x_{t}$    &0.0074  &0.003                   &0.2448  &0.115 \\
 $F_{e,1}$    &24.59  &11.2                   &200.9  &100.7 \\
 $F_{e,2}$    &27.75  &13.13                   &135.5  &56.5 \\
 $F_{e,3}$    &17.98  &4.61                   &222.1  &105.1 \\
 $F_{e,4}$    &30.70  &7.92                   &133.4  &55.7 \\
\cmidrule{1-5}\morecmidrules\cmidrule{1-5}
\end{tabular*}
\end{table}

The advantage of this proposed scheme in terms of tracking precision will be more competitive compared to its nonadaptive counterpart in which the pose feedback is also available when the velocity of the desired trajectory changes more sharply or the dynamic uncertainties are more severe.

	\section{Conclusion}\label{sec:conclusion}
This paper presents a fully distributed cooperation scheme for networked nonholonomic mobile manipulators in the presence of uncertain dynamics and interconnected kinematics. In the first phase of the scheme pipeline, an adaptation-based estimation law is established for each mobile manipulator to estimate the linearized desired trajectory of the virtual leader in a distributed way. Pose formation idea is incorporated here to achieve cooperative task allocation considering the offset between the Cartesian-space frame and EE frame of robotic agent. Then based on the existing works, physical constraints imposed on the desired load distribution and desired internal force are proposed in terms of the wrench synthesis, which lays a foundation for tight cooperation problem. In the last phase, a set of distributed adaptive control is proposed to achieve synchronization between the motion of the mobile manipulator ensemble irrespective of the kinematic and dynamic uncertainties in both the mobile manipulators and the tightly grasped object. The controlled synchronization between the EE motions of the networked NMMs contributes to the improvement of the cooperative task tracking performance and the transient performance quantified by 2-norm of the interaction/internal forces. In addition, redundancy of each robotic agent is locally resolved in the velocity level and is utilized to achieve subtasks based on multi-priority strategy. Noisy Cartesian-space velocities are totally avoided here. This complete scheme is independent from the object's center of mass by employing task-oriented strategy and formation-based cooperative control and does not require any persistent excitation condition to achieve the tracking objectives. It is theoretically proved that the convergence of the task tracking error and synchronization error are guaranteed. Finally, simulation results of two typical cooperation tasks, i.e. kinematic cooperation and dynamic cooperation, are both illustrated to validate the efficacy and demonstrate the expected performance of the proposed scheme.

	\appendix
\setcounter{table}{0}
\setcounter{equation}{0}
 \captionsetup[table]{font=footnotesize,labelsep=newline,skip=3pt,justification=centering}        
\renewcommand{\thetable}{A-\Roman{table}}
\renewcommand{\tablename}{TABLE}
\renewcommand{\theequation}{A\arabic{equation}}

	\subsection{Proof of Property 2}
	\begin{proof}
	\begin{displaymath}
	\begin{split}
	 &{{M_{r,i}} = \left[ {\begin{array}{*{20}{c}}
{H_{v,i}^T{M_{v,i}}{H_{v,i}}}&{H_{v,i}^T{M_{vm,i}}}\\
{{M_{mv,i}}{H_{v,i}}}&{{M_{m,i}}}
\end{array}} \right]}\\
	 &{{B_{r,i}}({q_i}) = \left[ {\begin{array}{*{20}{c}}
{H_{v,i}^T{B_{v,i}}}&0\\
0&{{B_{m,i}}}
\end{array}} \right]}\\
	 &{{C_{r,i}} = \left[ {\begin{array}{*{20}{c}}
{H_{v,i}^T{C_{v,i}}{H_{v,i}} + H_{v,i}^T{M_{v,i}}{{\dot H}_{v,i}}}&{H_{v,i}^T{C_{vm,i}}}\\
{{M_{v,i}}{{\dot H}_{v,i}} + {C_{mv,i}}{H_{v,i}}}&{{C_{m,i}}}
\end{array}} \right]}\\
	 &{{G_{r,i}} = {[G_{v,i}^T{H_{v,i}},G_{m,i}^T]^T}}
	 \end{split}
	\end{displaymath}
	
	Then, we have
	\begin{align}   
         \label{eq:A1}
\begin{split}
&{{\dot M}_{r,i}} - 2{C_{r,i}} = \frac{{d\left[ {\begin{array}{*{20}{c}}
{H_{v,i}^T{M_{v,i}}{H_{v,i}}}&{H_{v,i}^T{M_{vm,i}}}\\
{{M_{mv,i}}{H_{v,i}}}&{{M_{m,i}}}
\end{array}} \right]}}{{dt}}\\
 &- 2\left[ {\begin{array}{*{20}{c}}
{H_{v,i}^T{C_{v,i}}{H_{v,i}} + H_{v,i}^T{M_{v,i}}{{\dot H}_{v,i}}}&{H_{v,i}^T{C_{vm,i}}}\\
{{M_{v,i}}{{\dot H}_{v,i}} + {C_{mv,i}}{H_{v,i}}}&{{C_{m,i}}}
\end{array}} \right]\\
 &= {\left[ {\begin{array}{*{20}{c}}
{{H_{v,i}}}&0\\
0&I
\end{array}} \right]^T}[{M_i}({q_i}) - {C_i}({q_i},{{\dot q}_i})]\left[ {\begin{array}{*{20}{c}}
{{H_{v,i}}}&0\\
0&I
\end{array}} \right]
\end{split}
    \end{align} 
    
    Considering the definition of the synthesized inertial matrix ${{\rm{ }}{M_{s,i}}}$ and Coriolis/Centrifugal matrix ${{C_{s,i}}}$ given in~\eqref{eq:19}
 \begin{align}   
         \label{eq:A2}
% {\begin{array}{c}
\begin{split}
&{{\dot M}_{s,i}} - 2{C_{s,i}} - {{\dot \beta }_i}J_{\phi ,i}^T{M_o}{J_{\phi ,i}}\\
 =&{{\dot M}_{r,i}} + {{\dot \beta }_i}(t)J_{\phi ,i}^T{M_o}{J_{\phi ,i}} + {\beta _i}(t)\frac{{dJ_{\phi ,i}^T{M_o}{J_{\phi ,i}}}}{{dt}} \\
 &-2({C_{r,i}} + {\beta _i}(t)[J_{\phi ,i}^T{C_o}{J_{\phi ,i}} + J_{\phi ,i}^T{M_o}\frac{{d{J_{\phi ,i}}}}{{dt}}])\\
 &- {{\dot \beta }_i}J_{\phi ,i}^T{M_o}{J_{\phi ,i}}\\
 =&({{\dot M}_{r,i}} - 2{C_{r,i}}) + {\beta _i}(t)J_{\phi,i}^T({{\dot M}_o} - 2{C_o}){J_{\phi ,i}}\\
 =&{\left[ {\begin{array}{*{20}{c}}
{{H_{v,i}}}&0\\
0&I \end{array}} \right]^T}[{M_i}({q_i}) - {C_i}({q_i},{{\dot q}_i})]\left[ {\begin{array}{*{20}{c}}
{{H_{v,i}}}&0\\
0&I
\end{array}} \right]\\
&+ {\beta _i}(t)J_{\phi ,i}^T({{\dot M}_o} - 2{C_o}){J_{\phi,i}}
\end{split}
\end{align} 

	 \begin{table*}[ht]
\begin{equation}
\label{eq:AA3}
    \begin{split}
{{\dot V}_i} =&2e_i^T\left( {1 - \frac{1}{{{K_\varepsilon }}}} \right){K_{s,i}}{\Lambda _i}{{\dot e}_i} - s_i^T\left( {{K_{{\vartheta _i}}} - \frac{1}{2}{{\dot \beta }_i}J_{\phi ,i}^T{M_o}{J_{\phi ,i}}} \right){s_i} + \tilde \theta _{k,i}^T\Gamma _{k,i}^{ - 1}{{\dot{\tilde \theta}}_{k,i}} + \tilde x_{o,i}^T{\alpha _i}{K_{o,i}}{{\dot{\tilde x}}_{o,i}}\\
&- s_i^T\hat J_{e,i}^T{K_{s,i}}{{\hat J}_{e,i}}{s_i} + s_i^TJ_{\phi ,i}^T{{\tilde F}_{I,i}} - {k_{r,i}}{({{\hat J}_{e,i}}{s_i})^T}{{\hat J}_{e,i}}{s_i}\\
 \le&2e_i^T\left( {1 - \frac{1}{{{K_\varepsilon }}}} \right){K_{s,i}}{\Lambda _i}{{\dot e}_i} + \tilde \theta _{k,i}^T\Gamma _{ki}^{ - 1}{{\dot{\tilde \theta}}_{k,i}} + \tilde x_{o,i}^T{\alpha _i}{K_{o,i}}{{\dot{\tilde x}}_{o,i}} - s_i^T\left( {{K_{{\vartheta _i}}} - {\vartheta _i}} \right){s_i} + s_i^TJ_{\phi ,i}^T{{\tilde F}_{I,i}}\\
 &- {k_{r,i}}{\left\| {{{\hat J}_{e,i}}{s_i}} \right\|^2} - {\left( {{S_{x,i}} + {Y_{k,i}}{{\tilde \theta }_{k,i}} - {{\dot{\tilde x}}_{o,i}}} \right)^T}{K_{s,i}}\left( {{S_{x,i}} + {Y_{k,i}}{{\tilde \theta }_{k,i}} - {{\dot{\tilde x}}_{o,i}}} \right)\\
 \le&s_i^TJ_{\phi ,i}^T{{\tilde F}_{I,i}}- {\left( {{S_{x,i}} + {\alpha _i}{{\tilde x}_{o,i}}} \right)^T}{K_{s,i}}\left( {{S_{x,i}} + {\alpha _i}{{\tilde x}_{o,i}}} \right) - s_i^T\left( {{K_{{\vartheta _i}}} - {\vartheta _i}} \right){s_i} + \tilde x_{o,i}^T{\alpha _i}{K_{o,i}}\left( {{Y_{k,i}}({\zeta _i},{{\dot \zeta }_i}){{\tilde \theta }_{k,i}} - {\alpha _i}{{\tilde x}_{o,i}}} \right)\\
 &- {k_{r,i}}{\left\| {{{\hat J}_{e,i}}{s_i}} \right\|^2} + \tilde \theta _{k,i}^T\Gamma _{k,i}^{ - 1}{{\dot{\tilde \theta}}_{k,i}} + 2e_i^T\left( {1 - \frac{1}{{{K_\varepsilon }}}} \right){K_{s,i}}{\Lambda _i}{{\dot e}_i}\\
 \le&- \alpha _i^2\tilde x_{o,i}^T\left( {{K_{s,i}} + {K_{o,i}}} \right){{\tilde x}_{o,i}} - 2{\alpha _i}{K_{s,i}}S_{x,i}^T{{\tilde x}_{o,i}} - s_i^T\left( {{K_{{\vartheta _i}}} - {\vartheta _i}} \right){s_i} - S_{x,i}^T{K_{s,i}}{S_{x,i}} + s_i^TJ_{\phi ,i}^T{{\tilde F}_{I,i}}\\
 &- {k_{r,i}}{\left\| {{{\hat J}_{e,i}}{s_i}} \right\|^2} + 2e_i^T\left( {1 - \frac{1}{{{K_\varepsilon }}}} \right){K_{s,i}}{\Lambda _i}{{\dot e}_i}\\
 \le&- \left( {1 - \frac{1}{{{K_\varepsilon }}}} \right)S_{x,i}^T{K_{s,i}}{S_{x,i}} + 2e_i^T\left( {1 - \frac{1}{{{K_\varepsilon }}}} \right){K_{s,i}}{\Lambda _i}{{\dot e}_i} - s_i^T\left( {{K_{{\vartheta _i}}} - {\vartheta _i}} \right){s_i} - \alpha _i^2\left( {{K_{s,i}} + {K_{o,i}} - {K_\varepsilon }{K_{s,i}}} \right)\tilde x_{o,i}^T{{\tilde x}_{o,i}}\\
 &+ s_i^TJ_{\phi ,i}^T{{\tilde F}_{I,i}} - {k_{r,i}}{\left\| {{{\hat J}_{e,i}}{s_i}} \right\|^2}
\end{split}
\end{equation}

\end{table*}

    According to the property of the dynamics of mechanical system, one can know that ${({M_i} - {C_i})}$ and ${({\dot M_o} - 2{C_o})}$ are skew-symmetric matrices by particular choice of ${C_i}$ and ${C_o}$. Therefore, skew-symmetric property of ${{\dot M_{s,i}} - 2{C_{s,i}} - {\dot \beta _i}J_{\phi ,i}^T{M_o}{J_{\phi ,i}}}$ is proved.
	\end{proof}
	
	\subsection{Proof of Theorem 1}	
	\begin{proof}
	Before establishing the stability analysis, we would like to introduce a corollary and a lemma first. Based on the passive decomposition approach~\cite{li2010stability} and input-output property~\cite{ioannou1996robust}, the following corollary can be concluded:

	\emph{Corollary A-1.} Define $x = {[{x_1},{x_2}...{x_n}]^T}$ and $y = {[{y_1},{y_2}...{y_n}]^T}$. For the first-order dynamics $\dot x =  - {{\cal L}}x + y$ where ${{\cal L}}$ is the Laplacian matrix associated with a directed graph containing a spanning tree, $x \in {{{\cal L}}_p}$ and $\mathord{\buildrel{\lower3pt\hbox{$\scriptscriptstyle\frown$}} \over x}  \in {{{\cal L}}_p}$ if $y \in {{{\cal L}}_p}$ for $p \in \left[ {1,\infty } \right]$ with $\mathord{\buildrel{\lower3pt\hbox{$\scriptscriptstyle\frown$}} \over x}  = {[{x_1} - {x_2},{x_2} - {x_3}...,{x_{n - 1}} - {x_n}]^T}$ [36].

	\emph{Lemma A-1.} If $f,{\rm{ }}\dot f \in {{{\cal L}}_\infty }$ and $f \in {{{\cal L}}_p}$ for some ${p = [1,\infty )}$, then $f(t) \to 0$ as $t \to \infty $~\cite{ioannou1996robust}.

	The stability analysis should be decomposed into two steps: 
	
	Step 1. ${{\dot V_{d,o}} \le 0}$ from~\eqref{eq:55} implies that ${{V_{d,o}}}$, the orientation component of Lyapunov candidate ${{V_d}}$, defined in~\eqref{eq:51} is bounded and nonincreasing, which means ${{\varepsilon _{o,d}} \in {{{\cal L}}_\infty }}$, ${{\tilde \theta _{tr,o,i}} \in {{{\cal L}}_\infty }}$ and ${{\delta _o} \in {{{\cal L}}_2}}$. Considering that the desired trajectory ${{x_{td}}}$ and its time derivative are bounded, then we have ${{\hat o_{d,i}} \in {{{\cal L}}_\infty }}$ from~\eqref{eq:24} and further   from~\eqref{eq:25}. Together with the fact that ${{\gamma _o} = {[\gamma _1^T,...,\gamma _N^T]^T} \in {\Re ^{3N}}}$ relates to ${\delta _o}$ by ${\gamma _o} = \left( {{{\cal L}} + {{\cal B}}} \right){\delta _o}$, ${\gamma _o} \in {{{\cal L}}_\infty }$ and further ${{\dot{\hat o}_{d,i}} \in {{{\cal L}}_\infty }}$ with ${{\dot{\hat \theta}_{tr,o,i}} \in {{{\cal L}}_\infty }}$ can be obtained based on~\eqref{eq:28} and~\eqref{eq:29}. Hence ${{\dot \delta _{o,i}} \in {{{\cal L}}_\infty }}$ from the derivative of~\eqref{eq:25}. In addition, we have ${{\dot \gamma _{o,i}} \in {{{\cal L}}_\infty }}$ by differentiating~\eqref{eq:26}, then ${{\ddot{\hat \theta}_{tr,o,i}} \in {{{\cal L}}_\infty }}$, ${{\ddot{\hat o}_{d,i}} \in {{{\cal L}}_\infty }}$ and further ${{\ddot \delta _{o,i}} \in {{{\cal L}}_\infty }}$.
	
	\emph{Conclusion 1}: With the above analysis, ${{\delta _{o,i}} \in {{{\cal L}}_\infty } \cap {{{\cal L}}_2}}$, ${{\dot \delta _{o,i}} \in {{{\cal L}}_\infty }}$ and ${{\ddot \delta _{o,i}} \in {{{\cal L}}_\infty }}$ holds, then ${{\delta _{o,i}} \to 0}$ and ${{\dot \delta _{o,i}} \to 0}$ as $t \to \infty $ are achieved, $\forall i \in \{ 1,2,...,N\} $.
	
	Step 2: From~\eqref{eq:55}, ${{\dot V_{d,p}} = {{ - \kappa \delta _p^T\left( {{{\cal Q}} \otimes {I_3}} \right){\delta _p}} \mathord{\left/ {\vphantom {{ - \kappa \delta _p^T\left( {{{\cal Q}} \otimes {I_3}} \right){\delta _p}} 2}} \right. \kern-\nulldelimiterspace} 2} \le 0}$ when ${{\bar \rho _i} = 0}$, which means that closed-loop dynamics of position formation is input-to-state stable. Based on the results of Conclusion 1, ${{{\tilde {\cal T}}_{ji}},{{\dot{\tilde {\cal T}}}_{ji}},{{\tilde {\cal T}}_{ti}},{{\dot{\tilde {\cal T}}}_{ti}} \in {{{\cal L}}_\infty }}$ and ${{{\tilde {\cal T}}_{ji}},{{\dot{\tilde {\cal T}}}_{ji}},{{\tilde {\cal T}}_{ti}},{{\dot{\tilde {\cal T}}}_{ti}} \to 0}$ as $t \to \infty $, thus ${\mathop {\lim }\limits_{t \to \infty } {\bar \rho _i} = 0}$ holds. Then following the similar proof procedure in Step 1, one can easily obtain that ${{\delta _{p,i}} \to 0}$ and ${{\dot \delta _{p,i}} \to 0}$ as $t \to \infty $ based on its input-to-state stability property with bounded vanishing disturbance. Together with Conclusion 1, the following conclusion can be stated:
	
	\emph{Conclusion 2}: ${{\delta _i} \to 0}$ and ${{\dot \delta _i} \to 0}$ as $t \to \infty $ are achieved, $\forall i \in \{ 1,2,...,N\} $.
	\end{proof}

\subsection{Proof of Theorem 2}
	\begin{proof}
Detail of~\eqref{eq:63} is presented in~\eqref{eq:AA3}.
  ${\dot V_i} \le 0$ implies that Lyapunov candidate defined in~\eqref{eq:57} is always bounded and non-increasing, which immediately means ${s_i} \in {{{\cal L}}_2} \cap {{{\cal L}}_\infty }$, ${e_i} \in {{{\cal L}}_2} \cap {{{\cal L}}_\infty }$, ${\dot e_i} \in {{{\cal L}}_2}$, ${{\tilde \theta _{d,i}},{\rm{ }}{\tilde \theta _{k,i}},{\rm{ }}{\tilde \theta _{Id,i}} \in {{{\cal L}}_\infty }}$, ${\tilde x_{o,i}} \in {{{\cal L}}_2} \cap {{{\cal L}}_\infty }$.
  
  The collective form of~\eqref{eq:38} can be written as:
        \begin{align}   
         \label{eq:A3}
{\Delta {x_o} =  - \left( {{{\cal E}{\cal L}} \otimes {I_6}} \right)\int_0^t {\Delta {x_o}}  + e}
    \end{align}
    where ${\Delta {x_o}}$ and ${e}$ are the column stack vectors of ${\Delta {x_{o,i}}}$ and ${{e_i}}$. ${{{\cal E}} = diag\left( {{\varepsilon _i}} \right)}$ is a positive constant diagonal matrix. Then by using Corollary A-1, ${\Delta {x_{o,i}},{\rm{ }}\int_0^t {\Delta {x_{o,j}}}  \in {{{\cal L}}_2} \cap {{{\cal L}}_\infty }}$, ${\Delta {x_{o,i}} - \Delta {x_{o,j}} \in {{{\cal L}}_2} \cap {{{\cal L}}_\infty }}$, ${\int_0^t {\left( {\Delta {x_{o,i}} - \Delta {x_{o,j}}} \right)}  \in {{{\cal L}}_2} \cap {{{\cal L}}_\infty }}$ can be obtained from the above equation since ${e_i} \in {{{\cal L}}_2} \cap {{{\cal L}}_\infty }$. This further yield ${\int_0^t {\Delta {x_{o,i}}}  \to 0}$ and ${\int_0^t {\left( {\Delta {x_{o,i}} - \Delta {x_{o,j}}} \right)}  \to 0}$ according to Lemma A-1. Since ${{\hat \theta _{k,i}} \in {{{\cal L}}_\infty }}$, ${\hat J_{e,i}^\dag ({\zeta _i},{\hat \theta _{k,i}}) \in {{{\cal L}}_\infty }}$ is guaranteed with the incorporated singularity-robust technique~\eqref{eq:41}, ${{\dot \zeta _{r,i}} \in {{{\cal L}}_\infty }}$ can be concluded from~\eqref{eq:40} considering ${{e_i} \in {{{\cal L}}_\infty }}$ and boundedness of the estimated desired trajectories, i.e. ${\hat x_{d,i}},{\rm{ }}{\dot{\hat x}_{d,i}} \in {{{\cal L}}_\infty }$. Then from~\eqref{eq:43}, ${s_i} \in {{{\cal L}}_\infty }$ gives rise to ${{\dot \zeta _i} \in {{{\cal L}}_\infty }}$ and thus ${{\dot x_{e,i}} = {Y_{k,i}}({\zeta _i},{\dot \zeta _i}){\theta _{k,i}} \in {{{\cal L}}_\infty }}$. In addition, ${\dot{\tilde x} _{o,i}} \in {{{\cal L}}_\infty}$ can be easily obtained from~\eqref{eq:37} and ${{\dot{\tilde x}_{o,i}} = {\dot x_{o,i}} - {\dot x_{e,i}}}$ further indicates ${{\dot x_{o,i}}, \Delta {\dot x_{o,i}}, \Delta {\dot x_{o,i}} - \Delta {\dot x_{o,j}} \in {{{\cal L}}_\infty }}$. Together with ${\tilde x_{o,i}} \in {{{\cal L}}_2} \cap {{{\cal L}}_\infty }$, ${\tilde x_{o,i}} \to 0$ holds according to Lemma A-1. ${\Delta {x_{o,i}} \to 0}$, ${\Delta {x_{o,i}} - \Delta {x_{o,j}} \to 0}$ can be derived from Barbalat's lemma since ${\int_0^t {\Delta {x_{o,i}}}  \to 0}$ and ${\int_0^t {\left( {\Delta {x_{o,i}} - \Delta {x_{o,j}}} \right)}  \to 0}$. Based on the above analysis, ${\Delta {x_{e,i}} \to 0}$ and ${\Delta {x_{e,i}} - \Delta {x_{e,j}} \to 0}$ are guaranteed.
  
  With ${{\dot \zeta _i} \in {{{\cal L}}_\infty }}$ and ${\tilde x_{o,i}} \in {{{\cal L}}_\infty }$, ${{\dot{\hat \theta}_{k,i}} \in {{{\cal L}}_\infty }}$ can be obtained from~\eqref{eq:48}, which further gives rise to ${{\ddot \zeta _{r,i}} \in {{{\cal L}}_\infty }}$ considering that ${{\ddot{\hat x}_{d,i}} \in {{{\cal L}}_\infty }}$. Then from~\eqref{eq:46} and~\eqref{eq:34}, the boundedness of the commanded torques is guaranteed, i.e. ${{\tau _i} \in {{{\cal L}}_\infty }}$.
  
  The joint-space dynamics of the mobile manipulators~\eqref{eq:9} can be reformulated in its task space and the compact form of the constraint interconnected system's dynamics is given as:
        \begin{align}   
         \label{eq:A4}
{{\bar M_x}\ddot{\bar X }= {\bar F_\Sigma } + {\bar F_E}}
    \end{align}
   where ${{\bar M_x} = diag({M_o},{M_{x,1}},...,{M_{x,N}})}$ is the collective inertial matrix of the interconnected robotic system and ${{M_{x,i}}{\rm{ = }}{(J_{e,i}^\dag )^T}{M_{r,i}}J_{e,i}^\dag }$ denotes the Cartesian-space counterpart of ${{M_{r,i}}}$, ${\ddot{\bar X} = {[\ddot x_o^T,\ddot x_{e,1}^T,...,\ddot x_{e,N}^T]^T}}$ is the stack acceleration. ${{\bar F_\Sigma } = {[{( - {C_o}{\dot x_o} - {g_o})^T},\bar F_{x,1}^T,\bar F_{x,2}^T,...,\bar F_{x,N}^T]^T}}$ in which ${{\bar F_{x,i}} = {F_{x,i}} - {C_{x,i}}{\dot x_{e,i}} - {G_{x,i}}}$. ${{F_{x,i}} = {(J_{e,i}^\dag )^T}{B_{r,i}}{\tau _i}}$, ${{C_{x,i}}{\dot x_{e,i}} = {(J_{e,i}^\dag )^T}{C_{r,i}}{\dot \zeta _i} - {M_{x,1}}{\dot J_{e,i}}{\dot \zeta _i}}$ and ${{G_{x,i}} = {(J_{e,i}^\dag )^T}{G_{r,i}}}$ are Cartesian-space counterparts of the input joint torque ${{\tau _i}}$, the  Coriolis/Centrifugal force and the gravity ${{G_{r,i}}}$. ${{\bar F_E} = {[F_o^T, - F_{e,1}^T,..., - F_{e,N}^T]^T}}$ is the stack interaction force. 
   
   Based on the analysis in~\cite{sieber2018human}, the closed-form of the interaction force ${{\bar F_E}}$ in our case here can be given as
           \begin{align}   
         \label{eq:A5}
{{\bar F_E} = {A^T}{(A\bar M_x^{ - 1}{A^T})^{ - 1}}(b - A\bar M_x^{ - 1}{\bar F_\Sigma })}
    \end{align}
    where\\
    ${b = {[ {{{( {S{{({\omega _o})}^2}{r_1}} )}^T},0,{{( {S{{({\omega _o})}^2}{r_2}} )}^T},0,...,{{( {S{{({\omega _o})}^2}{r_N}} )}^T},0,} ]^T}}$ and ${A = [ - G_o^T,{I_{6N}}]}$.
    
    From the above equation, one can easily conclude that ${{\bar F_E} \in {{{\cal L}}_\infty }}$ which further leads to ${\ddot \zeta _i} \in {{{\cal L}}_\infty }$, ${\ddot x_{e,i}} \in {{{\cal L}}_\infty }$ and ${\dot s_i} \in {{{\cal L}}_\infty }$ considering~\eqref{eq:9} and~\eqref{eq:43}. Together with ${{\dot{\hat \theta}_{k,i}} \in {{{\cal L}}_\infty }}$, boundedness of ${{\hat{\ddot x}_{e,i}}}$ can be guaranteed since ${{\hat{\ddot x}_{e,i}} = {\dot{\hat J}_{e,i}}({\dot \zeta _i},{\dot{\hat \theta}_{k,i}}){\dot \zeta _i} + {\hat J_{e,i}}({\zeta _i},{\hat \theta _{k,i}}){\ddot \zeta _i}}$. Then from~\eqref{eq:35}, ${{\ddot x_{o,i}} \in {{{\cal L}}_\infty }}$ can be concluded. This yields ${{\ddot{\tilde x}_{o,i}} = {\ddot x_{o,i}} - {\ddot x_{e,i}} \in {{{\cal L}}_\infty }}$, which means that ${{\dot{\tilde x}_{o,i}}}$ is uniformly continuous. Then according to Barbalat's Lemma, ${\dot{\tilde x}_{o,i}} \to 0$ is achieved since ${\tilde x_{o,i}} \to 0$ has been proved. Furthermore, since the estimated desired trajectory is also bounded, i.e. ${\rm{ }}{\ddot{\hat x}_{d,i}} \in {{{\cal L}}_\infty }$, then ${{\rm{ }}\Delta {\ddot x_{o,i}} = {\ddot x_{o,i}} - {\ddot{\hat x}_{d,i}} \in {{{\cal L}}_\infty }}$ and further ${\Delta {\ddot x_{o,i}} - \Delta {\ddot x_{o,j}} \in {{{\cal L}}_\infty }}$ can be guaranteed considering ${{\ddot x_{o,i}} \in {{{\cal L}}_\infty }}$. Together with ${\Delta {x_{o,i}} \to 0}$ and ${\Delta {x_{o,i}} - \Delta {x_{o,j}} \to 0}$, ${\Delta {\dot x_{o,i}} \to 0}$ and ${\Delta {\dot x_{o,i}} - \Delta {\dot x_{o,j}} \to 0}$ can be obtained. Similar to the preceding procedure, ${\Delta {\dot x_{e,i}} \to 0}$ and ${\Delta {\dot x_{e,i}} - \Delta {\dot x_{e,j}} \to 0}$ as $t \to \infty $, $\forall i \in \{ 1,2,...,N\} $.
    
    Then we will give the analysis of internal force tracking error. 
               \begin{align}   
         \label{eq:A6}
\begin{split}
{J_{\phi ,i}}M_{s,i}^{ - 1}J_{\phi ,i}^T{{\tilde F}_{I,i}}&={J_{\phi ,i}}M_{s,i}^{ - 1}[({K_{{\vartheta _i}}} + {k_{r,i}}){s_i}+ {Y_{Id,i}}{{\tilde \theta }_{Id,i}}]\\
&+{J_{\phi ,i}}{{\dot s}_i} + {J_{\phi ,i}}M_{s,i}^{ - 1}( {\hat J_{e,i}^T{K_{s,i}}{{\hat J}_{e,i}} - {C_{s,i}}}){s_i}\\
&- {J_{\phi ,i}}M_{s,i}^{ - 1}{{Y_{d,i}}({\zeta _i},{{\dot \zeta }_i},{{\dot \zeta }_{r,i}},{{\ddot \zeta }_{r,i}},{\beta _i}){{\tilde \theta }_{d,i}}} 
\end{split}
    \end{align}
   with ${s_i},{\dot s_i} \in {{{\cal L}}_\infty }$ and ${{\tilde \theta _{d,i}},{\tilde \theta _{Id,i}} \in {{{\cal L}}_\infty }}$, the boundedness of ${{\tilde F_{I,i}}}$ is implied considering the positive definiteness of ${{J_{\phi ,i}}M_{s,i}^{ - 1}J_{\phi ,i}^T}$.
	\end{proof}	
	
	\subsection{Parameters and some fundamental matrices employed in the simulation}
	
\begin{table}[h]
\footnotesize{
\caption{PHYSICAL PARAMETERS OF THE MOBILE MANIPULATOR AND THE GRASPED OBJECT} 
\begin{tabular*}{0.48\textwidth}{c@{\extracolsep{\fill}}*{6}{c}}
\hline \hline
Part &Body  &${m_i(kg)}$  &${I_i(kg.m^2)}$ &${l_i(m)}$ &${l_{ci}(m)}$\\
\multirow{3}{*}{Manipulator} 
&Link1 & 6.5  &0.12   &0.4    &0.28\\
&Link2 & 5.0  &0.42   &0.285    &0.20\\
&Link3 & 2.6  &0.10   &0.35    &0.25\\
\hline

\multirow{2}{*}{Mobile base} &Mass & Inertial  &$r$   &$l_M$    &COM\\
&10 &${diag([0,0,1])}$  &0.15   &0.5    &[0,0,0]\\

\multirow{2}{*}{Object} &Mass & Inertial  &COM   &    & \\
&6 &${diag([0,0,8])}$  &[1,0,0]   &     &  \\

 \hline \hline
\end{tabular*}}
\end{table}
\begin{table}[h]
\footnotesize{
\caption{UNCERTAINTIES OF THE INITIAL PARAMETERS KNOWN TO EACH ROBOT} 
\begin{tabular*}{0.48\textwidth}{c@{\extracolsep{\fill}}*{4}{c}}
\cmidrule{1-5}\morecmidrules\cmidrule{1-5}
                     &Robot 1  &Robot 2      &Robot 3  &Robot4 \\
\midrule
 Trajectory    &0\%  &17.12\%          &32.94\%  &45.67\% \\
 Dynamics    &20\%  &25\%         &15\%  &20\% \\
 Kinematics    &10\%  &15\%           &15\%  &10\% \\
\cmidrule{1-5}\morecmidrules\cmidrule{1-5}
\end{tabular*}}
\end{table}

\begin{table}
\footnotesize{
\caption{PARAMETERS SELECTED FOR THE CONTROL SCHEME IN CASE 2} 
\begin{tabular*}{0.48\textwidth}{c@{\extracolsep{\fill}}*{3}{l}}
\hline \hline
\multirow{2}{*}{Task Allocation} &$\kappa$ & 2\\
 &${\Gamma _{tr,i}}$ &20 \\
\hline
\multirow{10}{*}{Task Allocation} &${\alpha _i}$ & 20\\
 &${\varepsilon _i}$ &30 \\
  &${\Lambda _i}$  &$10{I_3}$ \\
   &${\Delta _i}$ &0.05 \\
    &${\lambda _{\max i}}$ &0.1 \\
     &${K_{s,i}}$ &$5$ \\
      &${K_{{\vartheta _i}}}$ &$10$ \\
       &${\Gamma _{d,i}}$ &$0.001{I_3}$\\
        &${\Gamma _{k,i}}$ &$5{I_3}$\\
         &${K_{o,i}}$ &$20$\\
         \hline \hline
\end{tabular*}}
\end{table}

	The forward kinematics of the mobile manipulator is given as
	\begin{displaymath}
	 {\begin{array}{c}
{X_{e,i}} = {l_{1,i}}\cos ({\varphi _{v,i}} + {q_{m1,i}}) + {l_{2,i}}\cos ({\varphi _{v,i}} + {q_{m1,i}} + {q_{m2,i}})\\
 + {l_{3,i}}\cos ({\varphi _{v,i}} + {q_{m1,i}} + {q_{m2,i}} + {q_{m3,i}}) + {X_{W,i}} + {l_{M,i}}\cos \left( {{\varphi _{v,i}}} \right)
\end{array}}
	\end{displaymath}
	\begin{displaymath}
	 {\begin{array}{c}
{Y_{e,i}} = {l_{1,i}}\sin ({\varphi _{v,i}} + {q_{m1,i}}) + {l_{2i}}\sin ({\varphi _{v,i}} + {q_{m1,i}} + {q_{m2,i}})\\
 + {l_{3,i}}\sin ({\varphi _{v,i}} + {q_{m1,i}} + {q_{m2,i}} + {q_{m3,i}}) + {Y_{W,i}} + {l_{M,i}}\sin \left( {{\varphi _{v,i}}} \right)
\end{array}}
	\end{displaymath}
	\begin{displaymath}
	 {{\varphi _{e,i}} = {\varphi _{v,i}} + {q_{m1,i}} + {q_{m2,i}} + {q_{m3,i}}}
	\end{displaymath}
	where ${[{X_{e,i}},{Y_{e,i}},{\varphi _{e,i}}]^T}$ denotes the end-effector configuration of the mobile manipulator.
	
	The generalized coordinates and reduced coordinates of the mobile manipulator are defined as:
	\begin{flushleft}
	 ${q_i} = {[{x_{v,i}},{y_{v,i}},{\varphi _{v,i}},{q_{m1,i}},{q_{m2,i}},{q_{m3,i}}]^T}$\\
	 
	 ${\zeta _i} = {[{q_{R,i}},{q_{L,i}},{q_{m1,i}},{q_{m2,i}},{q_{m3,i}}]^T}$
	\end{flushleft}
	
	The input transformation matrix and relationship between two coordinates of the mobile manipulator are presented as:
	\begin{displaymath}
	 {{B_{r,i}} = \left[ {\begin{array}{*{20}{c}}
{{{\cos ({\varphi _{v,i}})} \mathord{\left/
 {\vphantom {{\cos ({\varphi _{v,i}})} r}} \right.
 \kern-\nulldelimiterspace} r}}&{{{\cos ({\varphi _{v,i}})} \mathord{\left/
 {\vphantom {{\cos ({\varphi _{v,i}})} r}} \right.
 \kern-\nulldelimiterspace} r}}&0&0&0\\
{{{\sin ({\varphi _{v,i}})} \mathord{\left/
 {\vphantom {{\sin ({\varphi _{v,i}})} r}} \right.
 \kern-\nulldelimiterspace} r}}&{{{\sin ({\varphi _{v,i}})} \mathord{\left/
 {\vphantom {{\sin ({\varphi _{v,i}})} r}} \right.
 \kern-\nulldelimiterspace} r}}&0&0&0\\
{{b \mathord{\left/
 {\vphantom {b {2r}}} \right.
 \kern-\nulldelimiterspace} {2r}}}&{{{ - b} \mathord{\left/
 {\vphantom {{ - b} {2r}}} \right.
 \kern-\nulldelimiterspace} {2r}}}&0&0&0\\
0&0&1&0&0\\
0&0&0&1&0\\
0&0&0&0&1
\end{array}} \right]}
	\end{displaymath}
	\begin{displaymath}
	 {{H_{v,i}} = \left[ {\begin{array}{*{20}{c}}
{{{r\cos ({\varphi _{v,i}})} \mathord{\left/
 {\vphantom {{r\cos ({\varphi _{v,i}})} 2}} \right.
 \kern-\nulldelimiterspace} 2}}&{{{r\cos ({\varphi _{v,i}})} \mathord{\left/
 {\vphantom {{r\cos ({\varphi _{v,i}})} 2}} \right.
 \kern-\nulldelimiterspace} 2}}\\
{{{r\sin ({\varphi _{v,i}})} \mathord{\left/
 {\vphantom {{r\sin ({\varphi _{v,i}})} 2}} \right.
 \kern-\nulldelimiterspace} 2}}&{{{r\sin ({\varphi _{v,i}})} \mathord{\left/
 {\vphantom {{r\sin ({\varphi _{v,i}})} 2}} \right.
 \kern-\nulldelimiterspace} 2}}\\
{{r \mathord{\left/
 {\vphantom {r b}} \right.
 \kern-\nulldelimiterspace} b}}&{{{ - r} \mathord{\left/
 {\vphantom {{ - r} b}} \right.
 \kern-\nulldelimiterspace} b}}
\end{array}} \right]}
	\end{displaymath}
	\begin{displaymath}
	 {{A_{v,i}} = \left[ {\sin \left( {{\varphi _{v,i}}} \right), - \cos \left( {{\varphi _{v,i}}} \right),0} \right]}
	\end{displaymath}
	\begin{displaymath}
	 {{\dot q_i} = \left[ {\begin{array}{*{20}{c}}
{\frac{{r\cos ({\varphi _{v,i}})}}{2}}&{\frac{{r\cos ({\varphi _{v,i}})}}{2}}&0&0&0\\
{\frac{{r\sin ({\varphi _{v,i}})}}{2}}&{\frac{{r\sin ({\varphi _{v,i}})}}{2}}&0&0&0\\
{\frac{r}{b}}&{ - \frac{r}{b}}&0&0&0\\
0&0&1&0&0\\
0&0&0&1&0\\
0&0&0&0&1
\end{array}} \right]\left[ {\begin{array}{*{20}{c}}
{{{\dot q}_{R,i}}}\\
{{{\dot q}_{L,i}}}\\
{{{\dot q}_{m1,i}}}\\
{{{\dot q}_{m2,i}}}\\
{{{\dot q}_{m3,i}}}
\end{array}} \right]}
	\end{displaymath}
	
\subsection{Nomenclature and abbreviation}
\begin{table}[h]
\footnotesize{
\caption{NOMENCLATURE} 
% \begin{tabular}{l}{\textwidth}
\begin{tabular}{lp{5.6cm}}
\midrule \midrule
Abbreviation       &Definition \\
\midrule
${\rm{\hat A}}$            &Estimated form of the related matrix/vector A\\
${\rm{\tilde A}}$            &Parameter estimation error of the related matrix/vector A\\
${q_i}$            &Generalized coordinates of \emph{i}th mobile manipulator\\
${\zeta _i}$            &Reduced coordinates of \emph{i}th mobile manipulator\\
${x_{e,i}}$            &End-effector pose vector of \emph{i}th mobile manipulator\\
${{J_{e,i}}}$           &Reduced Jacobian matrix of \emph{i}th mobile manipulator\\
${{Y_{k,i}}({\zeta _i},{\dot \zeta _i})}$           &Kinematic regressor matrix\\
${{\theta _{k,i}}}$           &Linearized kinematic parameters\\
${x_t}$            &Task-space coordinates at the object's operational point\\
${x_{td}}$            &Desired cooperative trajectory\\
${Y_{d,i}}({\zeta _i},{\dot \zeta _i},{\ddot \zeta _i},{\beta _i})$            &Dynamic regressor matrix\\
${\theta _{d,i}}$            &Linearized dynamic parameters\\
${M_{s,i}}$            &Synthesized inertial matrix of the \emph{i}th mobile manipulator\\
${C_{s,i}}{\dot \zeta _i}$            &Synthesized Coriolis and centrifugal forces of the \emph{i}th mobile manipulator\\
${G_{s,i}}$          &Gravitational forces of the \emph{i}th mobile manipulator\\
${{{\cal F}}_{tr}}\left( t \right)$          &Collective basis function of the desired cooperative task\\
${\theta _{tr}}$          &Linearized parameter of the desired cooperative task\\
${\delta _i}$          &Allocated task estimation error defined in~\eqref{eq:25}\\
${\gamma _i}$          &Consensus error defined in~\eqref{eq:26}\\
${{{\cal T}}_{ji}}$          &Relative displacement and orientation between EE coordinates of \emph{i}th and \emph{j}th robot\\
${{{\cal T}}_{ti}}$          &Relative displacement and orientation between operational point and end-effector coordinate of \emph{i}th robot\\
${\dot x_{o,i}}$          &Observed EE velocity defined in~\eqref{eq:35} \\
${e_i}$          &Cross-coupling error defined in~\eqref{eq:38} \\
${S_{x,i}}$         &Cartesian-space sliding variable defined in~\eqref{eq:39} \\
${\dot \zeta _{r,i}}$          &Reference joint velocity defined in~\eqref{eq:40} \\
${\tilde x_{o,i}} = {x_{o,i}} - {x_{e,i}}$          &Observer error of the \emph{i}th mobile manipulator \\
$\Delta {x_{e,i}} = {x_{e,i}} - {\hat x_{d,i}}$          &Tracking error of the \emph{i}th mobile manipulator\\
$\Delta {x_{o,i}} = {x_{o,i}} - {\hat x_{d,i}}$          &Error between the observed EE pose and desired EE trajectory\\
${\hat x_{d,i}}$          &Estimated local desired EE trajectory of the \emph{i}th mobile manipulator\\
${s_i}$          &Joint-space sliding vectors defined in~\eqref{eq:43}\\
${F_{e,i}}$          &External wrenches exerted by the holonomic constraint on EE of the \emph{i}th mobile manipulator\\
${F_{I,i}}$          &Internal wrench of the \emph{i}th mobile manipulatorr\\
${F_{Id,i}} = {[f_{Id,i}^T,\tau _{Id,i}^T]^T}$          &Desired internal wrench\\
${\varepsilon _{d,i}} = {[\varepsilon _{pd,i}^T,\varepsilon _{od,i}^T]^T}$          &Intermediate error variable defined in~\eqref{eq:53}\\
${\tau _i} = {[\tau _{v,i}^T,\tau _{m,i}^T]^T}$          &Input torques of the \emph{i}th mobile manipulator\\
\midrule \midrule
\end{tabular}}
\end{table}

\begin{table}[h]
\footnotesize
\caption{ABBREVIATION} 
\begin{tabular*}{0.48\textwidth}{l@{\extracolsep{\fill}}*{2}{l}}
\midrule \midrule
Abbreviation       &Definition \\
\midrule
COM            &Center of mass\\
DA            &Distributed adaptive control scheme proposed in this paper\\
DTVL            &Desired trajectory of the virtual leader\\
EE            &End-effector\\
ECCT            &Equivalent centroid of the kinematic cooperation task\\
IW            &Internal wrench\\
MW            &Motion-induced wrench\\
NA            &Conventional visual servoing control scheme without adaptation\\
NMM            &Nonholonomic mobile manipulator\\
RMS            &Root mean square\\
\midrule \midrule
\end{tabular*}
\end{table}

\section*{Acknowledgment}
The authors would like to express their appreciation to Hanlei Wang for his great help and so many insightful suggestions in this work. They also would like to thank S. Sosnowski and S. Music for their discussion on this work.

	\bibliographystyle{IEEEtran}
	\bibliography{FullyDistributedCooperation.bib}

\end{document}